\documentclass[journal]{IEEEtran}

\ifCLASSINFOpdf
  \usepackage[pdftex]{graphicx}
  \graphicspath{{../pdf/}{../jpeg/}}
  \DeclareGraphicsExtensions{.pdf,.jpeg,.png}
\fi

\UseRawInputEncoding

\usepackage[noadjust]{cite}  
\usepackage{multirow}
\usepackage{makecell}
\usepackage{booktabs}
\usepackage{threeparttable}
\usepackage{amsmath}
\usepackage{amssymb}
\usepackage{svg}
\usepackage{multicol}

\ifCLASSOPTIONcompsoc
    \usepackage[caption=false, font=normalsize, labelfont=sf, textfont=sf]{subfig}
\else
    \usepackage[caption=false, font=footnotesize]{subfig}
\fi

\usepackage[colorlinks, citecolor=blue, urlcolor=blue, bookmarks=false, hypertexnames=true]{hyperref}

\begin{document}

    






\title{\textcolor{black}{Vehicle-to-Everything Cooperative Perception for Autonomous Driving}}

\author{Tao Huang\IEEEauthorrefmark{1},~\IEEEmembership{Senior Member,~IEEE,}
        Jianan Liu\IEEEauthorrefmark{1},~
        Xi Zhou\IEEEauthorrefmark{1},~~\IEEEmembership{Graduate Student Member,~IEEE,}\\
        Dinh C. Nguyen,~\IEEEmembership{Member,~IEEE,}
        Mostafa Rahimi Azghadi,~\IEEEmembership{Senior Member,~IEEE,}
        Yuxuan Xia,~\IEEEmembership{Member,~IEEE,}\\
        Qing-Long Han\IEEEauthorrefmark{2},~\IEEEmembership{Fellow,~IEEE,} 
        and Sumei Sun,~\IEEEmembership{Fellow,~IEEE}
        
\thanks{\IEEEauthorrefmark{1}Equal contribution.}

\thanks{\IEEEauthorrefmark{2}Corresponding author.}

\thanks{T.~Huang, X.~Zhou, and M. R. Azghadi are with the College of Science and Engineering, James Cook University, Cairns, QLD 4870, Australia. Email: tao.huang1@jcu.edu.au, xi.zhou@jcu.edu.au, mostafa.rahimiazghadi@jcu.edu.au.}

\thanks{J.~Liu is with Momoni AI, Gothenburg, 40017, Sweden. Email: jianan.liu@momoniai.org.}

\thanks{Dinh C. Nguyen is with the Department of Electrical and Computer Engineering, University of Alabama in Huntsville, Huntsville, AL 35899, USA. Email: dinh.nguyen@uah.edu.}

\thanks{Yuxuan Xia is with the Department of Automation and Intelligent Sensing, Shanghai Jiaotong University, Shanghai 200240, China. Email: yuxuan.xia@sjtu.edu.cn.}

\thanks{Q.-L.~Han is with the School of Engineering, Swinburne University of Technology, Melbourne, VIC 3122, Australia. Email: qhan@swin.edu.au.}

\thanks{S.~Sun is with the Institute for Infocomm Research, Agency for Science, Technology and Research (A*STAR), Singapore 138632. Email: sunsm@i2r.a-star.edu.sg.}

\thanks{\textcolor{blue}{This article has been accepted for publication in \textit{\textbf{Proceedings of the IEEE}}.}} 

\thanks{\textcolor{blue}{DOI: \href{https://doi.org/10.1109/JPROC.2025.3600903}{10.1109/JPROC.2025.3600903}}}

\thanks{\textcolor{blue}{\textcopyright\ 2025 IEEE.  Personal use of this material is permitted.  Permission from IEEE must be obtained for all other uses, in any current or future media, including reprinting/republishing this material for advertising or promotional purposes, creating new collective works, for resale or redistribution to servers or lists, or reuse of any copyrighted component of this work in other works.}}

}

\markboth{Proceedings of the IEEE, Accepted in August~2025}
{\MakeLowercase{\textit{et al.}}: Demo of IEEEtran.cls for IEEE Journals}

\maketitle

\begin{abstract}

Achieving fully autonomous driving with enhanced safety and efficiency relies on vehicle-to-everything cooperative perception, which enables vehicles to share perception data, thereby enhancing situational awareness and overcoming the limitations of the sensing ability of individual vehicles. 
Vehicle-to-everything cooperative perception plays a crucial role in extending the perception range, increasing detection accuracy, and supporting more robust decision-making and control in complex environments.
This paper provides a comprehensive survey of recent developments in vehicle-to-everything cooperative perception, introducing mathematical models that characterize the perception process under different collaboration strategies. Key techniques for enabling reliable perception sharing, such as agent selection, data alignment, and feature fusion, are examined in detail.
In addition, major challenges are discussed, including differences in agents and models, uncertainty in perception outputs, and the impact of communication constraints such as transmission delay and data loss. 
The paper concludes by outlining promising research directions, including privacy-preserving artificial intelligence methods, collaborative intelligence, and integrated sensing frameworks to support future advancements in vehicle-to-everything cooperative perception.

\end{abstract}


\newpage
\section{Introduction}

\IEEEPARstart{C}ooperative \textcolor{black}{Perception (CP), enabled by Vehicle-to-Everything (V2X) communication \cite{8967260}, has emerged as a promising paradigm for enhancing situational awareness in autonomous driving. By enabling vehicles to exchange perception data with surrounding vehicles, infrastructure, pedestrians, and networks, CP extends visibility beyond line-of-sight limitations and enhances decision-making in complex scenarios, such as intersections and lane changes. This collaborative approach addresses the fundamental limitations of single-vehicle perception, which depends solely on onboard sensors such as cameras, LiDAR, and radar. 
}

\textcolor{black}{
In a full-stack autonomous driving system, core components, including perception, localization, planning, control, and system integration, must operate cohesively. V2X-based CP enhances each of these layers by enabling the exchange of sensory and positional data among connected agents\footnote{“Agents” refer to vehicles and infrastructure equipped with sensing and communication capabilities.}. At the perception level, shared information allows the detection of objects beyond a vehicle’s sensing range, improving overall accuracy. In localization, data from neighboring agents complements Global Positioning System (GPS) signals, particularly in areas with poor satellite coverage. For planning, real-time awareness of nearby trajectories enables more adaptive and informed decisions. These improvements also benefit the control layer, where enhanced situational awareness leads to more precise execution. Finally, system integration is strengthened through reduced reliance on individual sensors, enhancing the platform’s robustness and scalability. Collectively, V2X CP improves the intelligence and resilience of autonomous systems, supporting their safe deployment in real-world scenarios.
}

\begin{figure*}[t]
    \centering
    \subfloat[{Without V2X CP, each vehicle has a limited perception range.}]{
        \includegraphics[width=0.98\columnwidth]{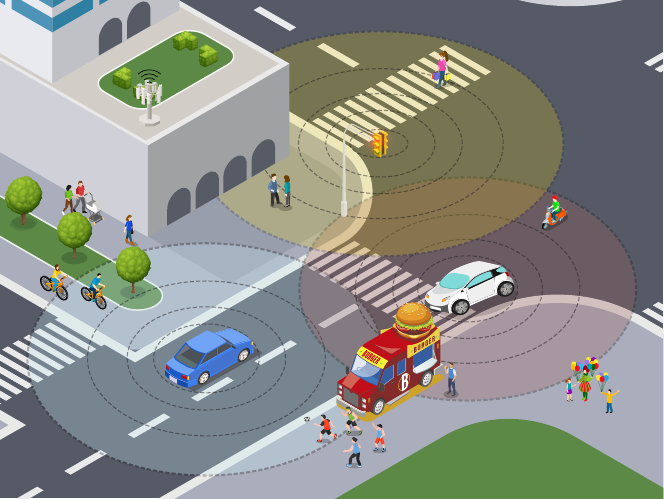} %
        \label{intro_NoCP} 
    }
    \hfill %
    \subfloat[With V2X CP, each vehicle gains an expanded awareness of its perception range.]{
        \includegraphics[width=0.98\columnwidth]{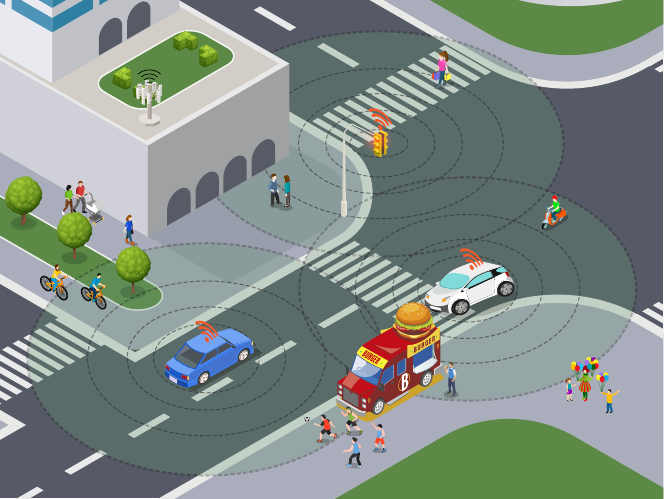} %
        \label{intro_WithCP}
    }
    \caption{An urban application scenario of V2X CP. The circles with different colours indicate the perception range, which is illustrative, as real sensing capabilities depend on factors such as sensor type and weather conditions.} 
    \label{intro}
\end{figure*}

\begin{figure*}[t]
    \centering
    \subfloat[Without V2X CP, the white car has a blocked front view and thus potentially can have a fatal collision with the running boy.]{
        \includegraphics[width=0.98\columnwidth]{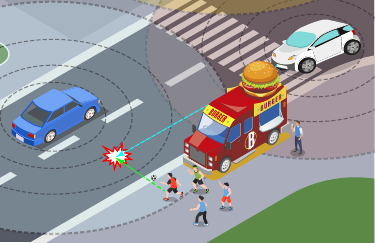}
        \label{intro_NoCP_}
    } 
    \hfill %
    \subfloat[With V2X CP, the white car can detect the running boy due to the expanded perception range, enabling it to predict the boy's behavior and avoid a potentially fatal collision.]{
        \includegraphics[width=0.98\columnwidth]{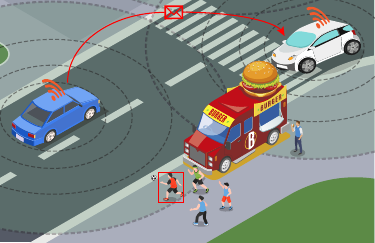}
        \label{intro_WithCP_}
    } 
    \caption{\textcolor{black}{An example of V2X CP improves road safety.}}
    \label{CP_Safety_example}
\end{figure*}

\textcolor{black}{
Fig. \ref{intro} illustrates an urban scenario comparing autonomous driving with and without CP. In Fig. \ref{intro_NoCP}, each vehicle operates independently. Each vehicle's limited sensing range, combined with obstructions, leads to unreliable detection of distant or occluded objects \cite{pilz2021components}. In contrast, Fig. \ref{intro_WithCP} depicts vehicles and infrastructure exchanging information via V2X communication, resulting in a more comprehensive understanding of the environment and the elimination of blind spots.
A specific case is shown in Fig. \ref{CP_Safety_example}, where a white hatchback's front view is blocked by a parked food truck, preventing it from detecting a child running into the street. This scenario, also shown in Fig. \ref{intro_NoCP_}, poses a serious collision risk if the vehicle cannot respond in time. With CP enabled, the blue Connected Autonomous Vehicle (CAV) communicates the child’s movement to the white hatchback, allowing it to decelerate proactively and avoid an accident, as illustrated in Fig. \ref{intro_WithCP_}.
}

\textcolor{black}{
While CP enhances driving safety, its practical deployment requires addressing several foundational challenges, including data privacy \cite{9583667}, ethical concerns \cite{9044647}, integration with smart cities and urban infrastructure \cite{10236573}, and regulatory and policy frameworks \cite{9001049}. Realizing the full potential of V2X CP, therefore, demands a coordinated, multi-dimensional effort.
}

\begin{table*}[t]
\scriptsize
\renewcommand{\arraystretch}{1.2}
\caption{A Summary of the Recently Published Surveys in Cooperative Perception with V2X Technologies.}\label{tab_I}
\centering
\renewcommand\tabcolsep{2.5pt}
\begin{threeparttable}
\begin{tabular}{cccccccccccccccccccccc}
\toprule
\multirow{2.5}{*}{Paper} &\multirow{2.5}{*}{Year} &\multirow{2.5}{*}{Publication} &\multirow{2.5}{*}{Survey Topic} &\multicolumn{18}{c}{The Taxonomy of Cooperative Perception System: Modules and Issues} \\
\cmidrule(l{2pt}r{1pt}){5-22}
&&&&CAS &PIS &PIC &DSRC &C-V2X &LC &CD &APE &CIF &MH &DH &DS &DP &PU &TD &S2R &Sim &Data \\
\midrule
\cite{yang2021machine} &2021 &\makecell[c]{IEEE\\Network} &\makecell[{{m{4.5cm}}}]{Feature level based cooperative object detection.} &$\times$ &$\surd$ &$\surd$ &$\times$ &$\times$ &$\times$ &$\surd$ &$\times$ &$\times$ &$\times$ &$\times$ &$\times$ &$\times$ &$\times$ &$\times$ &$\times$ &$\times$ &$\times$  \\
\cmidrule{1-22}
\cite{bai2022infrastructure} &2022 &\makecell[c]{IEEE\\IV} &\makecell[{{m{4.5cm}}}]{V2I cooperative object detection and tracking.} 
&$\times$ &$\times$ &$\times$ &$\times$ &$\times$ &$\times$ &$\times$ &$\times$ &$\surd$ &$\times$ &$\times$ &$\times$ &$\times$ &$\times$ &$\times$ &$\times$ &$\surd$ &$\surd$ \\
\cmidrule{1-22}
\cite{caillot2022survey} &2022 &\makecell[c]{IEEE\\TITS} &\makecell[{{m{4.5cm}}}]{Cooperation issues in localization, map generation, object detection and tracking.} 
&$\times$ &$\times$ &$\times$ &$\times$ &$\times$ &$\times$ &$\surd$ &$\surd$ &$\surd$ &$\times$ &$\times$ &$\times$ &$\times$ &$\times$ &$\times$ &$\times$ &$\surd$ &$\surd$ \\
\cmidrule{1-22}
\cite{xiang2023multi} &2023 &\makecell[c]{IEEE\\MITS} &\makecell[{{m{4.5cm}}}]{Multi-sensor fusion and CP.}  
&$\times$ &$\times$ &$\surd$ &$\times$ &$\times$ &$\times$ &$\surd$ &$\surd$ &$\surd$ &$\times$ &$\times$ &$\times$ &$\times$ &$\times$ &$\times$ &$\times$ &$\times$ &$\surd$ \\
\cmidrule{1-22}
\cite{bai2024survey} &\textcolor{black}{2024} &\makecell[c]{IEEE\\TITS} &\makecell[{{m{4.5cm}}}]{\textcolor{black}{Fundamental components and core aspects of a CP system for enabling cooperative driving automation.}} 
&$\times$ &$\times$ &$\surd$ &$\times$ &$\times$ &$\times$ &$\surd$ &$\surd$ &$\surd$ &$\times$ &$\times$ &$\times$ &$\times$ &$\times$ &$\times$ &$\times$ &$\surd$ &$\surd$ \\
\cmidrule{1-22}
\cite{han2023collaborative} &2023 &\makecell[c]{IEEE\\MITS} &\makecell[{{m{4.5cm}}}]{V2X cooperation efficiency, robustness, and safety.} 
&$\times$ &$\surd$ &$\surd$ &$\times$ &$\times$ &$\times$ &$\surd$ &$\surd$ &$\surd$ &$\surd$ &$\times$ &$\surd$ &$\times$ &$\times$ &$\surd$ &$\times$ &$\times$ &$\surd$ \\
\cmidrule{1-22}
\cite{survey_runsheng_jiaqi} &\textcolor{black}{2023} &$-$ &\makecell[{{m{4.5cm}}}]{\textcolor{black}{The challenges of implementing CP within V2X systems and the practical applicability of different CP methodologies in V2X environments.}} 
&$\surd$ &$\surd$ &$\surd$ &$\times$ &$\times$ &$\surd$ &$\surd$ &$\surd$ &$\surd$ &$\surd$ &$\surd$ &$\surd$ &$\surd$ &$\times$ &$\times$ &$\surd$ &$\surd$ &$\surd$ \\
\cmidrule{1-22}
\cite{cp_survey_tiv} &\textcolor{black}{2024} &\makecell[c]{IEEE\\TIV} &\makecell[{{m{4.5cm}}}]{\textcolor{black}{CP in intelligent transportation systems, with emphasizing road-to-vehicles collaboration at intersections.}} &$\times$ &$\surd$ &$\surd$ &$\times$ &$\times$ &$\times$ &$\surd$ &$\surd$ &$\surd$ &$\times$ &$\times$ &$\surd$ &$\surd$ &$\times$ &$\times$ &$\times$ &$\times$ &$\surd$ \\
\cmidrule{1-22}
\cite{10700687} &\textcolor{black}{2024} &\makecell[c]{IEEE\\TITS} &\makecell[{{m{4.5cm}}}]{\textcolor{black}{The perception architecture and associated technologies within the Vehicle-Road-Cloud Integration System.}} &$\times$ &$\surd$ &$\times$ &$\surd$ &$\surd$ &$\times$ &$\surd$ &$\times$ &$\surd$ &$\times$ &$\times$ &$\times$ &$\times$ &$\surd$ &$\times$ &$\times$ &$\times$ &$\times$\\
\cmidrule{1-22}
\textbf{\makecell[c]{This\\survey}} &2025 &$-$ & \makecell[{{m{4.5cm}}}]{The evolution of V2X CP technologies, key elements of V2X CP, and V2X CP solutions that tackle major issues in practical driving scenarios.} &$\surd$ &$\surd$ &$\surd$ &$\surd$ &$\surd$ &$\surd$ &$\surd$ &$\surd$ &$\surd$ &$\surd$ &$\surd$ &$\surd$ &$\surd$ &$\surd$ &$\surd$ &$\surd$ &$\surd$ &$\surd$\\ 
\bottomrule
\end{tabular}
\begin{tablenotes}
\scriptsize  
\item[] \textbf{Note:} CAS: Cooperative Agent Selection, PIS: Perception Information Selection, PIC: Perception Information Compression, DSRC: Dedicated Short-range Communication, C-V2X: Cellular-based Vehicle-to-Everything, LC: Lossy Communication, CD: Communication Delays, APE: Alignment for Pose Errors, CIF: Cooperative Information Fusion, MH: Model Heterogeneity, DH: Data Heterogeneity, DS: Data Security, DP: Data Privacy, PU: Perception Uncertainties, TD: Task Discrepancy, S2R: Simulation to Reality, Sim: Simulator, Data: Dataset. The exact meaning of these items in this taxonomy will be explained systematically in the following sections.
\end{tablenotes}
\end{threeparttable}
\end{table*}

Although various approaches have been proposed to address these challenges, a comprehensive survey is needed to assess recent advances, identify research gaps, and propose future directions. Some existing surveys often focus on specific aspects of CP, such as feature-level cooperative object detection \cite{yang2021machine} or infrastructure-based CP \cite{bai2022infrastructure}, but lack a systematic overview of V2X CP technologies. To fill this gap, the work in \cite{caillot2022survey} provided an overview of CP algorithms, emphasizing challenges like localization and detection, while the review in \cite{xiang2023multi} discussed communication delays and information fusion in multi-sensor Vehicle-to-Vehicle (V2V) and Vehicle-to-Infrastructure (V2I) schemes. The survey in \cite{bai2024survey} explored CP system architectures and proposed a hierarchical framework for unifying CP scenarios. Despite covering common CP challenges, these surveys lacked an analysis of CP progress regarding practical implementation challenges.

Recent work \cite{han2023collaborative} evaluated collaborative modules and solutions focusing on efficiency, robustness, and safety. Another survey \cite{survey_runsheng_jiaqi} addressed practical challenges in implementing CP, such as latency, bandwidth constraints, and robustness, categorizing methodologies into early, intermediate, and late fusion. Environmental simulations were also conducted to assess mainstream CP approaches. A more recent survey \cite{cp_survey_tiv} focused on CP at intersections, analyzing perception, communication, and downstream applications. However, these surveys overlook the critical role of V2X communication, which is essential for enabling effective CP in autonomous vehicles by facilitating information transfer and supporting decision-making.
The work in \cite{10700687} examined typical V2X communication technologies, including Dedicated Short-range Communication (DSRC) and Cellular-based V2X communication (C-V2X), and presented an overview of perception architectures and technologies within the Vehicle-Road-Cloud Integration System, encompassing single-node, multi-node, and vehicle-road-cloud cooperative perception frameworks. While it offers insights into these technologies, it does not explore specific algorithms and techniques in depth. This work contributes to a comprehensive understanding of CP and its enabling V2X technologies and builds upon existing survey efforts. Table \ref{tab_I} compares our paper with these surveys. The key contributions of this paper are as follows:


\begin{itemize}

\item Filling gaps in existing surveys by conducting a thorough review of the evolution of V2X CP from its inception to state-of-the-art proposals.

\item Discussing established V2X communication technologies and the key impacts of real-world communication constraints on CP.

\item Presenting a unified framework for V2X CP to elucidate essential components of multi-agent collaboration, enhancing researchers' systematic understanding of the CP system.

\item Introducing a new taxonomy to review V2X CP solutions, addressing significant practical driving scenario issues.

\item Summarizing lessons learned, providing insights into existing challenges, and highlighting potential avenues for future research.

\end{itemize}

The remainder of this paper is organized as follows: Section \ref{math_formulat} presents the mathematical formulation of V2X CP. A brief historical overview of CP system development is provided in Section \ref{development_of_cooperative_perception_system}. Section \ref{modern_generic_framwork} introduces a generic framework for systematically understanding the CP system. Practical issues related to CP systems and relevant datasets are discussed in Sections \ref{issue_of_cooperative_perception} and \ref{dataset}, respectively. Section \ref{Performance_Validation} compares the performance of various CP methods in offline testing and explores real-world implementations. Section \ref{challenges} explores the challenges and potential future directions, and the paper concludes with Section \ref{conclusions}.

\section{\textcolor{black}{Mathematical Formulation of Cooperative Perception}} \label{math_formulat}

\begin{figure}[!t]
    \centering
    \subfloat[Early]{
        \includegraphics[width=0.28\columnwidth]{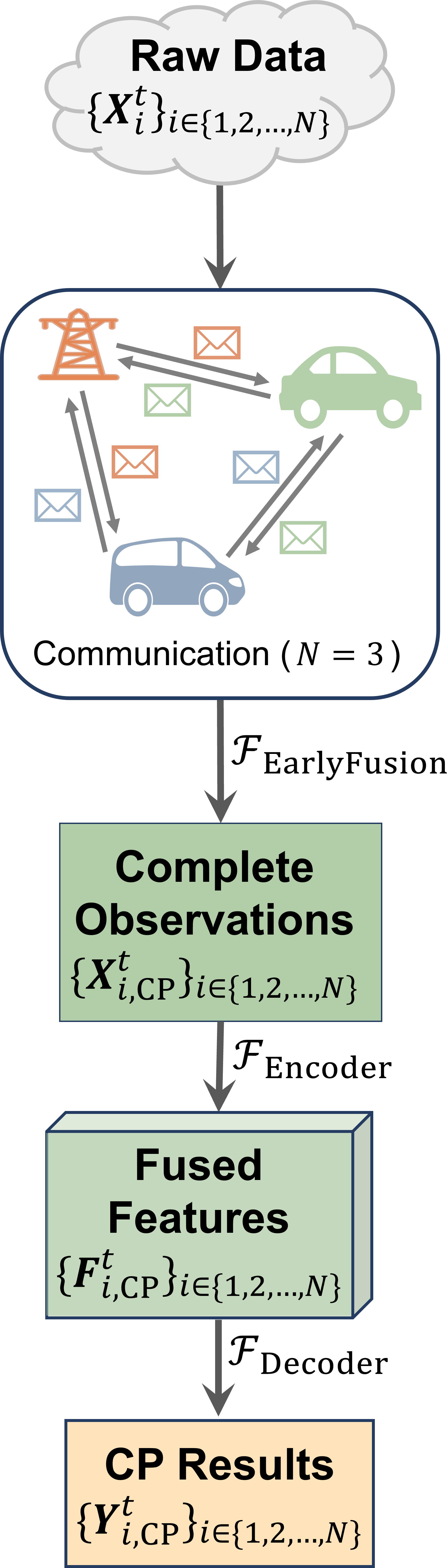}
        \label{earlyco}
    }
    \hfill 
    \subfloat[Intermediate]{
        \includegraphics[width=0.275\columnwidth]{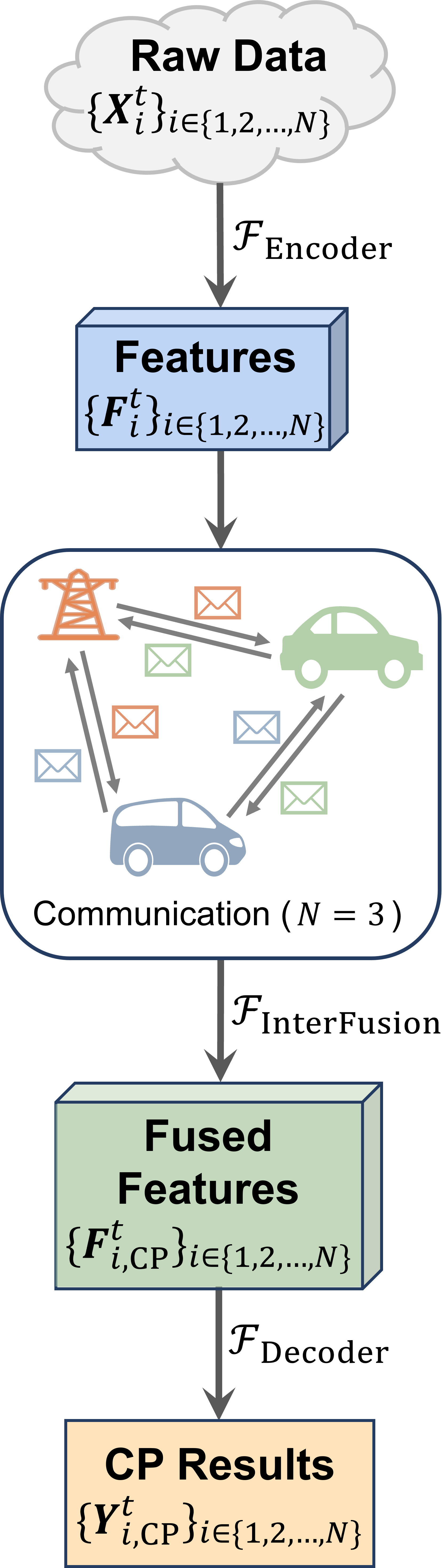}
        \label{lateco}
    }
    \hfill 
    \subfloat[Late]{
        \includegraphics[width=0.28\columnwidth]{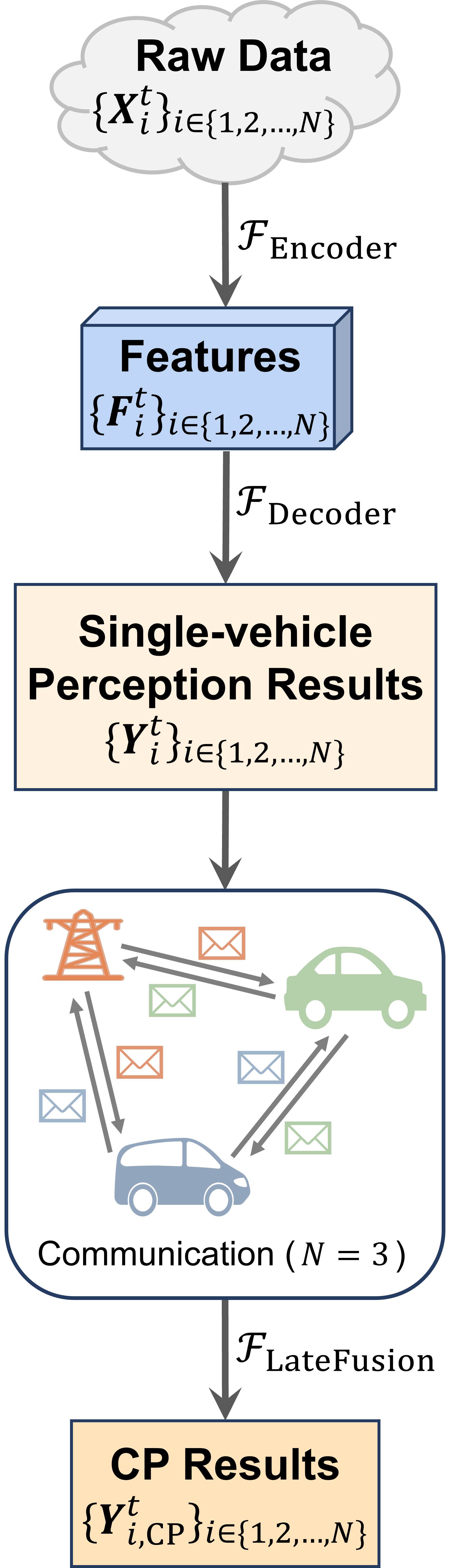}
        \label{intermediateco}
    }
    \caption{Overview of collaboration schemes for autonomous driving: (a) Early collaboration; (b) Intermediate collaboration; (c) Late collaboration.}
    \label{Co}
\end{figure}

\textcolor{black}{
CP strategies can be categorized into three types based on the processing stage at which data is shared: early, intermediate, and late collaboration. Let $\boldsymbol{X}_i^t$ denote the raw sensor data from agent $i$ at timestep $t$. Each agent $i$ uses an encoder, $\mathcal{F}_{\mathrm{Encoder}}(\cdot)$, to extract features $\boldsymbol{F}_i^t$ from raw sensor data, and then a decoder, $\mathcal{F}_{\mathrm{Decoder}} (\cdot)$, to generate perception results $\boldsymbol{Y}_i^t$. The key distinction across strategies lies in the data sharing stage and corresponding fusion method.
A schematic overview of these collaboration schemes is provided in Fig. \ref{Co}.
}

\subsection{\textcolor{black}{Early Collaboration}}
\textcolor{black}{
In early collaboration, agents exchange raw sensor data. 
Each agent $i$ fuses data from all $N$ agents:
\begin{equation}
\boldsymbol{X}_{i,\mathrm{CP}}^t = \mathcal{F}_{\mathrm{EarlyFusion} }\left ({\left \{ \boldsymbol{X}_k^t \right \} }_{k=1}^N   \right ).
\end{equation}
The fused raw data $\boldsymbol{X}_{i,\mathrm{CP}}^t$ is then encoded and decoded at agent $i$ to produce the final CP output. 
While this method provides the richest information, it imposes high bandwidth and synchronization requirements, making it less practical in communication-constrained environments.
}

\subsection{\textcolor{black}{Intermediate Collaboration}}
\textcolor{black}{
Intermediate collaboration involves sharing features extracted from raw data. 
Each agent $i$ encodes its observation as a feature $\boldsymbol{F}_i^t$ and exchanges it with peers. 
Fusion is performed on the feature level:
\begin{equation}
\boldsymbol{F}_{i,\mathrm{CP}}^t = \mathcal{F}_{\mathrm{InterFusion} }\left ({\left \{ \boldsymbol{F}_k^t \right \} }_{k=1}^N   \right ).
\end{equation}
The fused feature information $\boldsymbol{F}_{i,\mathrm{CP}}^t$ is then decoded at agent $i$ to obtain the CP output. 
This strategy balances communication efficiency and information richness, but its performance depends on the quality and consistency of features across agents.
}

\subsection{\textcolor{black}{Late Collaboration}}
\textcolor{black}{
In late collaboration, each agent $i$ completes local perception independently and shares only the final output $\boldsymbol{Y}_i^t$ with other agents.
These results are aggregated using a late fusion function:
\begin{equation}
\boldsymbol{Y}_{i,\mathrm{CP}}^t = \mathcal{F}_{\mathrm{LateFusion} }\left ({\left \{ \boldsymbol{Y}_k^t \right \} }_{k=1}^N   \right ). 
\end{equation}
This approach minimizes bandwidth requirements but lacks flexibility and detail, making it less suitable for dynamic or occluded environments.
}

\section{Development of Cooperative Perception: From Early Time Exploration till Recent Research}
\label{development_of_cooperative_perception_system}

\begin{figure*}[!t]
\centering
\includegraphics[width=1\textwidth]{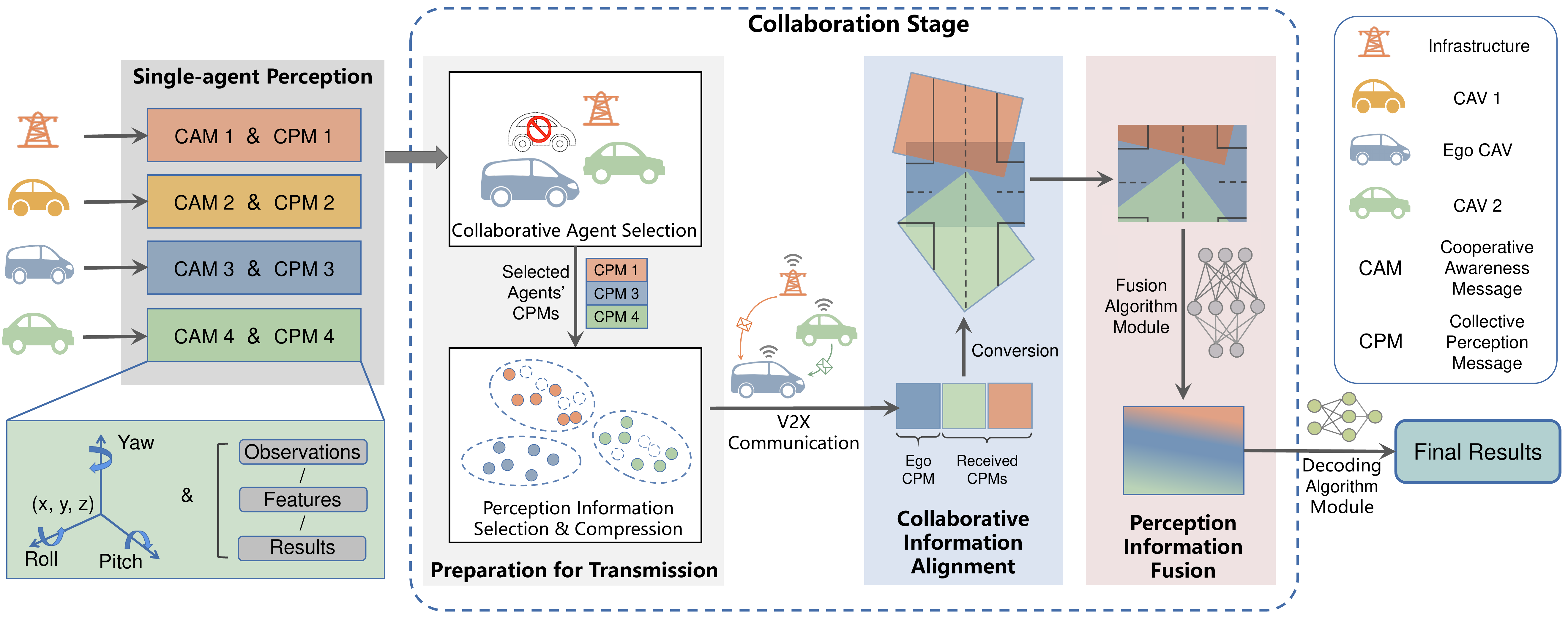}
\caption{The modern generic framework of the cooperative perception for autonomous driving.}
\label{CP_Process}
\end{figure*}

\textcolor{black}{
Early research in CP focused on rule-based and statistical signal processing methods, often relying on radar- or LiDAR-generated data or artificial sources. These approaches employed classical tracking and motion models (e.g., particle filters, Kalman filters, GM-PHD filters) to align and fuse object-level data from multiple vehicles \cite{rockl2008v2v, li2011multi, car2x2012rauch, C-GM-PHD2015vasic}. While these methods demonstrated engineering viability and improved perception coverage, they were limited by rigid motion assumptions, sensor noise sensitivity, and lack of scalability to real-world traffic dynamics. Additional efforts incorporated biologically inspired algorithms, such as mirror-neuron models for intention awareness \cite{kim2016cooperative}, though these approaches remained dependent on accurate pose estimation and often struggled in dynamic, multi-agent environments.
}

\textcolor{black}{
As CP evolved, limitations in extracting semantically rich and generalizable representations from shared data became apparent. This led to growing interest in deep learning, which has transformed single-vehicle perception and inspired learning-based CP frameworks to address traditional shortcomings. For example, cooperative object detection and classification tasks now benefit from deep architectures like YOLO and DenseNet to process shared data more effectively \cite{redmon2016you, huang2017densely, ml2018Rawashdeh}. Recent developments in this direction are discussed further in Sections \ref{modern_generic_framwork} and \ref{issue_of_cooperative_perception}.
}

\textcolor{black}{
Historically, CP systems have relied on late collaboration due to feature extraction complexity and communication constraints. With advances in deep learning, intermediate collaboration is gaining traction by enabling efficient feature sharing among agents \cite{dao2023practical}. However, this approach introduces challenges such as information redundancy and the need for robust data alignment and fusion algorithms. Optimizing the relevance of shared features and selecting appropriate collaborators are critical for effective multi-agent cooperation. Consequently, developing unified CP frameworks that balance communication efficiency, perception accuracy, and scalability is key to advancing Cooperative Driving Automation (CDA).
}

\section{Modern Generic Framework of Cooperative Perception} \label{modern_generic_framwork}

This section introduces a contemporary and comprehensive framework for CP in autonomous driving, along with relevant issues and methodologies.

\subsection{Framework Overview}
Given multiple agents capable of perception and communication within a shared environment, the objective is to enhance each agent's perceptual accuracy through distributed cooperation. Fig. \ref{CP_Process} illustrates the operational process of the proposed framework, comprising two major stages: single-agent perception and multi-agent collaboration.

During the single-agent perception phase, each agent acquires measurements of its Six Degrees-of-Freedom pose while simultaneously perceiving the surrounding environment. Communication messages are generated by packaging Cooperative Awareness Messages (CAMs) \cite{etsi2019cams} and Collective Perception Messages (CPMs) \cite{etsi2019cpms}. CAM messages convey the agent's status, encompassing information such as position and motion state, while CPM messages provide additional details regarding objects detected through sensor data, such as other vehicles and pedestrians. It is worth noting that the content of perceptual information may encompass raw sensor observations, intermediate features, and/or detection outcomes, contingent upon the chosen style of cooperation.

The collaboration stage encompasses four primary components, as depicted in Fig. \ref{CP_Process}: preparation for transmission, V2X communication, cooperative information alignment, and perception information fusion. To ensure efficient communication while preserving the accuracy of the cooperation system, the initial step involves carefully selecting suitable cooperative agents and pertinent perceptual information. Subsequently, each agent shares the CAM and the refined CPM with its peers through V2X communication \cite{V2AIX}. Following this, the ego vehicle aligns the acquired perception information with local data through coordinate conversion or pose transformation techniques. Finally, the aggregated data undergoes consolidation via a fusion algorithm, followed by decoding, resulting in a conclusive perception outcome. 
Note that all CAVs can transmit their perceptual data to a roadside infrastructure via V2X networks. The associated edge computing infrastructure is responsible for aggregating and consolidating this incoming data, which is then distributed to CAVs according to their specific needs.
In the subsequent sections, we provide detailed insights and a review of existing works on the collaboration stage's primary components.

\begin{table*}[!t]
\scriptsize
\renewcommand{\arraystretch}{1.3}
\caption{Summary of the Methods of Preparation for Transmission.}\label{tab_a}
\centering
\renewcommand\tabcolsep{3.5pt}
\begin{threeparttable}
\begin{tabular}{clccccccc}
\toprule  
Category &Method &Year &Scenario &Modality &Dataset/Simulator &Cooperation Style &Downstream Task &Available Code\\
\midrule 
\multirow{9}{*}{CAS} &\textcolor{black}{Who2com} \cite{liu2020who2com} &2020 &$-$ &Camera &AirSim-CP &Intermediate &Segmentation &\href{https://github.com/coperception/coperception}{$\surd$} \\
&\textcolor{black}{When2com} \cite{liu2020when2com} &2020 &$-$ &Camera &AirSim-MAP &Intermediate &Segmentation &\href{https://github.com/GT-RIPL/MultiAgentPerception}{$\surd$} \\
&CPOB-LearnCom \cite{wang2022collaborative} &2022 &V2I &LiDAR &CARLA-3D &Intermediate &Detection &$\times$\\
&AVUCB \cite{Jia2022online} &2022 &V2X &Camera &MS-COCO &Early &Detection &$\times$ \\
&MASS \cite{Jia2023MASSMS} &2023 &V2V &LiDAR &DOLPHINS &Early &Detection &$\times$ \\
&SelectComm \cite{chiu2023selective} &2023 &V2V &LiDAR &AUTOCASTSIM &Intermediate &Detection &$\times$ \\
&IoSI-CP \cite{liu2023rethinking} &2023 &V2X &LiDAR &OPV2V,V2XSet &Intermediate &Detection &\href{https://github.com/huangqzj/IoSI-CP/}{$\surd$} \\
&\textcolor{black}{InterCoop} \cite{InterCoop} &2024 &V2V &LiDAR &CARLA &Intermediate &Decision-making &$\times$\\
&\textcolor{black}{PACP} \cite{10646529} &2024 &V2V &Camera &OPV2V &Intermediate &Detection &$\times$\\
\cmidrule{1-9}
\multirow{16}{*}{PIS}  
&DyDisse \cite{allig2019dynamic} &2019 &V2X &Camera/LiDAR &$-$ &Late &Tracking &$\times$\\
&VehIdenti \cite{masuda2022feature} &2023 &V2X &$-$ &CARTERY &$-$ &$-$ &$\times$ \\
&RedunMitig \cite{redundancy2020} &2019 &V2X &$-$ &SUMO &Late &Detection &$\times$\\
&Look-ahead \cite{generation2020} &2019 &V2X &$-$ &SUMO &Late &Detection &$\times$\\
&AICP \cite{zhou2022aicp} &2022 &V2V &Camera &$-$ &Late &Detection,Tracking &\href{https://github.com/pengyuan-zhou/AICP}{$\surd$} \\
&DFS \cite{Bai2023Cooperverse} &2023 &V2X &LiDAR &CARTI &\makecell[c]{Early\&\\Intermediate \& Late} &Detection &$\times$\\
&\textcolor{black}{BM2CP} \cite{BM2CP} &2023 &V2V &LiDAR \& Camera &OPV2V, DAIR-V2X &Intermediate &Detection &\href{https://github.com/byzhaoAI/BM2CP}{$\surd$} \\
&DRLCP \cite{drlcp2020} &2020 &V2X &Camera &CIVS &Late &$-$ &$\times$\\
&FPV-RCNN \cite{keypoints2022yuan} &2022 &V2V &LiDAR &COMAP &Intermediate &Detection &\href{https://github.com/YuanYunshuang/FPV_RCNN}{$\surd$} \\
&UMC \cite{wang2023umc} &2023 &V2X &LiDAR &V2X-Sim,OPV2V &Intermediate &Detection &\href{https://github.com/ispc-lab/UMC}{$\surd$} \\
&GevBEV \cite{Yuan2023GeneratingEB} &2023 &V2V &LiDAR &OPV2V &Late &\makecell[c]{Detection,\\Segmentation} &$\times$ \\
&\textcolor{black}{IFTR} \cite{IFTR} &2024 &V2X &Camera &\makecell[c]{DAIR-V2X, V2XSet,\\OPV2V} &Intermediate &Detection &\href{https://github.com/wangsh0111/IFTR}{$\surd$} \\
&\textcolor{black}{CodeFilling} \cite{CP_Codebook} &2024 &V2X &Camera \& LiDAR &DAIR-V2X, OPV2VH+ &Intermediate &Detection &\href{https://github.com/PhyllisH/CodeFilling}{$\surd$} \\
&\textcolor{black}{V2X-PC} \cite{V2X-PC} &2024 &V2X &LiDAR &DAIR-V2X, V2XSet &Early &Detection &$\times$ \\
&\textcolor{black}{PAE} \cite{PillarAttention} &2024 &V2X &LiDAR &CARTI &Intermediate &Detection &$\times$ \\
&\textcolor{black}{ActFormer} \cite{ActFormer} &2024 &V2X &Camera &V2X-Sim &Intermediate &Detection &\href{https://coperception.github.io/ActFormer/}{$\surd$}  \\
\cmidrule{1-9}
\multirow{1}{*}{AIS}  
&Where2comm \cite{hu2022where2comm} &2022 &V2X &Camera/LiDAR &\makecell[c]{V2X-Sim,DAIR-V2X,\\OPV2V,CoPerception-UAVs} &Intermediate &Detection &\href{https://github.com/MediaBrain-SJTU/Where2comm}{$\surd$} \\
\cmidrule{1-9}
\multirow{12}{*}{PIC}
&V2VNet \cite{v2vnet2020wang} &2020 &V2V &LiDAR &V2V-Sim &Intermediate &\makecell[c]{Detection,\\Motion Prediction} &$\times$ \\
&\textcolor{black}{CODS} \cite{MultimodalCoop3DOD} &2023 &V2V &Camera \& LiDAR &CARLA &Intermediate &Detection &$\times$ \\
&\textcolor{black}{CPSC} \cite{CP_based_on_ST_feature} &2024 &V2X &LiDAR &DAIR-V2X, OPV2V &Intermediate &Detection &$\times$ \\
&FS-COD \cite{marvasti2020cooperative} &2020 &V2V &LiDAR &CARLA &Intermediate &Detection &$\times$ \\
&DiscoNet \cite{disconet2021li} &2021 &V2V &LiDAR  &V2X-Sim &Intermediate &Detection  &\href{https://github.com/ai4ce/DiscoNet}{$\surd$} \\
&V2X-ViT \cite{v2xvit2022xu} &2022 &V2X &LiDAR &V2XSet &Intermediate &Detection &\href{https://github.com/DerrickXuNu/v2x-vit}{$\surd$}\\
&COOPERNAUT \cite{cui2022coopernaut} &2022 &V2V &LiDAR &AUTOCASTSIM &Intermediate &Policy Learning &\href{https://github.com/UT-Austin-RPL/Coopernaut}{$\surd$} \\
&\textcolor{black}{CenterCoop} \cite{CenterCoop} &2024 &V2I &LiDAR &DAIR-V2X &Intermediate &Detection &$\times$\\
&Slim-FCP \cite{guo2022SlimFCP} &2022 &V2V &LiDAR &KITTI, T\&J &Intermediate &Detection &$\times$ \\
&\textcolor{black}{EMIFF} \cite{EMIFF} &2024 &V2X &Camera &DAIR-V2X &Intermediate &Detection &$\times$  \\
&\textcolor{black}{What2comm} \cite{yang2023what2comm} &2023 &V2X &LiDAR &\makecell[c]{DAIR-V2X,V2XSet,\\OPV2V} &Intermediate &Detection &$\times$ \\
&\textcolor{black}{SmartCooper} \cite{zhang2024smartcooper} &2024 &V2V &Camera &OpenCOOD &Intermediate &Detection &$\times$ \\
\bottomrule
\end{tabular}
\begin{tablenotes}
\scriptsize  
\item[] \textbf{Note:} CAS: Cooperative Agent Selection, PIS: Perception Information Selection, AIS: Agent and Information Selection, PIC: Perception Information Compression.
\end{tablenotes}
\end{threeparttable}
\end{table*}

\subsection{Preparation for Transmission}
In CP systems, continuous information exchange among multiple agents is crucial for achieving comprehensive traffic awareness. However, real-world application scenarios often face limitations in communication bandwidth. Excessive network load increases the risk of important data packets being delayed or lost, which could compromise the performance and safety of CP systems. Therefore, it is essential to balance the trade-off between perception performance and communication efficiency. To address this challenge, researchers have proposed various strategies aimed at reducing the volume of shared data prior to transmission and alleviating the burden on vehicular networks. As summarized in Table \ref{tab_a}, these strategies can be grouped into four main approaches:
\begin{itemize}
\item Cooperative Agent Selection: This approach focuses on minimizing communication costs by transmitting perception data through a partially connected agent graph, thereby reducing redundancy within the network.
\item Perception Information Selection: \textcolor{black}{This category aims to select a subset of perceptive data critical to the ego vehicle's situational awareness. It focuses on maximizing informativeness, which refers to the quality of information, specifically the extent to which the data is relevant, valuable, and reliable in enhancing perception accuracy.} 
\item Agent and Information Selection: This category encompasses a joint consideration of both cooperative agent selection and perception information selection, aiming to balance communication efficiency and data importance.
\item \textcolor{black}{Perception Information Compression: This strategy involves compressing or encoding essential information before sharing, ensuring that the transmission rate remains within the required data rate limits.}
\end{itemize}

\subsubsection{Cooperative Agent Selection}
Selecting the right collaborators is crucial because inappropriate candidates can cause disruptions and disturbances that could harm overall performance. 

The attention mechanism, recognized for its efficacy in improving model efficiency by focusing on relevant data, offers promise in selecting suitable agents for the ego CAV. In this context, a multi-stage handshake communication mechanism was proposed in Who2com \cite{liu2020who2com}, employing the general attention mechanism to select the appropriate collaborators. This mechanism computes matching scores among agents, selectively engaging the most pertinent agents to optimize bandwidth usage. 
Building upon this, a scaled general attention mechanism was introduced in When2com \cite{liu2020when2com} to determine optimal communication timing. Through this mechanism, communication was initiated by the ego vehicle only when its own information was deemed insufficient, thereby allowing network resources to be conserved.
Although this multi-stage communication process proves effective, its implementation and optimization in dynamically varying environments pose significant challenges. 
To address this, a dynamic communication mechanism was proposed in \cite{wang2022collaborative}, allowing vehicles to determine, based on learned attention weights, which infrastructure nodes to engage with, thereby balancing detection accuracy and communication resource utilization.

To minimize energy consumption, the data sharing scheduling problem can be framed as a variant of the Multi-Armed Bandit (MAB) problem \cite{slivkins2019introduction}, factoring in perception performance, time-varying wireless channels, and power consumption. In this regard, an online learning-based algorithm known as the Adaptive Volatile Upper Confidence Bound (AVUCB) \cite{Jia2022online} was proposed to schedule the most advantageous vehicle, offering an enhanced view while adhering to communication bandwidth constraints. However, it does not account for the high mobility of cooperative vehicles. To address this issue, the Mobility-Aware Sensor Scheduling (MASS) algorithm \cite{Jia2023MASSMS} formulated decentralized CP scheduling as a Restless-MAB (RMAB) problem \cite{whittle1988restless}. In the RMAB context, rewards continually evolve, requiring each vehicle to continuously learn about its surroundings while selecting actions that empirically maximize rewards. When applied to a practical SUMO trace, the MASS algorithm demonstrated the ability to enhance overall perception gain without incurring additional costs associated with frequent meta-information exchanges.

The selection of cooperative agents can also be likened to the Multi-agent Path Finding (MAPF) problem, in which paths are planned for multiple agents \cite{ma2021learning}. Inspired by MAPF, a selective communication algorithm was proposed in \cite{chiu2023selective} to determine a subset of cooperative vehicles based on estimated information gains from connected vehicles. In this approach, information gains were estimated by the ego vehicle through comparison between its own detected object positions and the received two-dimensional center positions from other connected vehicles. Vehicles associated with the highest estimated information gains were then selected to define the communication scope. In contrast to random selection strategies, this algorithm demonstrated higher success rates while incurring minimal additional communication costs.

\textcolor{black}{
Addressing the need for prioritization in CP, the Priority-Aware Collaborative Perception (PACP) framework leverages a Bird’s Eye View (BEV) based matching strategy to determine the relative importance of nearby CAVs based on their spatial correlation with the ego vehicle \cite{10646529}.
}

All the aforementioned methods primarily focused on spatial correlations, overlooking the significance of temporal dependencies in collaborator selection. To address this limitation, IoSI-CP \cite{liu2023rethinking} was proposed to exploit semantic information across both temporal and spatial dimensions for optimal collaborator selection. This approach leveraged a Graph Neural Network (GNN) to capture inherent correlations between the ego vehicle and its neighbouring agents. Agents with weights of zero or lower were considered ineligible for collaboration, which eliminates unsuitable collaborators responsible for noise. Nevertheless, this approach relied on one-dimensional weights that may not fully capture the intricacies addressed by high-dimensional representation methods. To enhance collaborator relevance, InterCoop \cite{InterCoop} was developed to compute an interaction score for each vehicle by analyzing geometric properties of the road network in conjunction with the temporal dynamics of trajectories, thereby prioritizing vehicles most relevant to the prevailing driving conditions.

\subsubsection{Perception Information Selection}
The selection of targeted information for the ego vehicle is another essential factor in efficient CP. Recently, a growing focus has been on the selection of cooperative message content to improve communication efficiency. A number of methods have emerged for selecting or generating CPMs intended for transmission. These methods can be broadly categorized into rule-based, distance-based, and learning-based approaches, each rooted in distinct mechanisms and theoretical principles.

The rule-based approach to CPM generation/selection entails designing rules that dictate which aspects of detected object information should be included in a CPM. 
For example, in \cite{allig2019dynamic}, a rule was proposed in which information about an object’s mode is transmitted only if the object is turning or accelerating. Similarly, in \cite{masuda2022feature}, visual information from detected vehicles was used to determine their communication identifiers, which were then employed to integrate CPMs with local perception data. 
All detected vehicles that have exchanged CAM messages will not be considered in future CPMs. Standardization efforts, led by the European Telecommunications Standards Institute (ETSI), have also introduced CPM generation rules, specifying what information CPMs should include and when to generate new messages. 
In \cite{redundancy2020} and \cite{generation2020}, improved algorithms were proposed to refine these rules by filtering out redundant information about detected objects that had been recently transmitted by nearby vehicles. Although rule-based methods are relatively straightforward to implement, they primarily rely on changes in the position and speed of detected objects.

Distance-based methods prioritize data based on distance metrics, considering the spatial location relative to the ego vehicle's onboard sensors to measure data point significance. For instance, the Augmented Informative Collaborative Perception (AICP) system \cite{zhou2022aicp} employed a fitness sorting algorithm based on Mahalanobis distance to select crucial information at the application level, effectively filtering out less relevant packets with minimal delay. 
In a separate approach, Dynamic Feature Sharing (DFS) \cite{Bai2023Cooperverse} was developed to identify relevant feature data for sharing using Random-K priorities based on Manhattan distance. Within this framework, early, intermediate, and late fusion strategies were incorporated to optimize dynamic node engagement in the cooperative perception system.

While distance-based methods emphasize the transmission of feature cells or data points close to the ego vehicle, they may only partially meet the ego vehicles' perception information requirements in real-world scenarios. In contrast, learning-based approaches leverage the entirety of perception data to intelligently determine information selection for sharing. In BM2CP \cite{BM2CP}, a modality-guided collaboration was designed to selectively share the most critical multi-modal features across agents, thereby promoting communication efficiency. This was achieved by generating a preference threshold mask to filter BEV features. Deep Reinforcement Learning (DRL) allows agents to learn policies through interactions with their environment, maximizing cumulative rewards \cite{9945653}. The work \cite{drlcp2020} introduced a DRL-based Cooperative Perception (DRLCP) scheme, where each CAV intelligently decides what information to transmit based on the context captured by its onboard sensors. DRLCP's primary objective is to determine whether to transmit or discard data, but it was evaluated on only two road networks without considering the impact of obstructions on vehicular communications.

Another category of learning-based approaches, the selective-region mechanism, focuses on selecting CPM information within appropriate communication regions. For instance, in \cite{keypoints2022yuan}, a bounding box matching module was employed, and strategic critical point selection was adopted to choose features enclosed within bounding box proposals for transmission. In contrast, UMC \cite{wang2023umc} was proposed with a trainable region-wise communication selection module, Entropy-CS, in which entropy was leveraged to distinguish informative regions and select suitable areas for transmission based on resolution levels. GevBEV \cite{Yuan2023GeneratingEB} was developed to use evidential BEV maps for identifying areas where the ego vehicle required additional information, and evidential deep learning with Dirichlet evidence \cite{sensoy2018evidential} was applied to quantify point-based classification uncertainty, thereby highlighting the most relevant regions for communication.

Foreground regions contain more informative content than background regions.
Based on this principle, a message selection and feature map reconstruction module was employed in IFTR \cite{IFTR} to transmit the most relevant visual features from foreground regions.
In CodeFilling \cite{CP_Codebook}, the information-filling-driven message selection was designed to determine the ego vehicle's perceptual needs by solving a local optimization problem based on the information score map gathered from other agents. 
An optimized selection matrix was generated to determine which spatial regions should be prioritized for transmission.
In spatial selection approaches, only the informative regions identified by spatial confidence maps are transmitted; however, under strict bandwidth constraints, this may result in the loss of critical object information \cite{V2X-PC}.
To address this issue, a Point Cluster Packing module was introduced in V2X-PC \cite{V2X-PC}, in which important points were sampled to control bandwidth usage. This point cluster representation was shown to preserve object features during message packing, support efficient message aggregation regardless of collaboration range, and enable the explicit communication of object structural information.

Attention mechanism enables a single agent to intelligently identify the most relevant perception information from multiple agents. For example, in the pillar attention encoder (PAE) \cite{PillarAttention}, attention values were extracted for each feature to serve as importance indicators, and the most significant features were selected for transmission based on available communication bandwidth. In ActFormer \cite{ActFormer}, an active selection network was proposed to assign each query with an interest score, which indicates the relevance and significance of each feature for the ego's perception. This method actively selects the most relevant sensory data from multiple robots based on spatial knowledge without relying on the sensory measurements themselves, which improves the efficiency of multi-agent object detection.

\subsubsection{Agent and Information Selection}

Methods that consider both cooperative agents and information selection jointly can offer a more balanced approach to optimizing precision and bandwidth in CP.
One such method, Where2comm \cite{hu2022where2comm}, was introduced as a spatial-confidence-aware cooperative perception framework in which spatial confidence maps are generated by each agent to identify perceptually critical regions in the feature map. 
These critical features were compactly packaged into messages and shared via a sparsely connected communication graph. A communication graph construction strategy was employed to selectively determine communication partners, thereby minimizing unnecessary bandwidth usage. Furthermore, only non-zero features and their corresponding indices were transmitted. 
Experimental results have demonstrated that Where2comm enables mutually beneficial cooperation by transmitting spatially sparse yet crucial features through vehicle-to-everything communication, thereby optimizing both perception accuracy and communication efficiency.

\subsubsection{Perception Information Compression}

Compression techniques have been shown to effectively prioritize and retain the most critical features of perception data, ensuring that essential information required for perception tasks is preserved even when data size is reduced. For example, in V2VNet \cite{v2vnet2020wang}, a variational image compression algorithm \cite{variational_image_compression} was employed to compress feature maps. In \cite{MultimodalCoop3DOD}, feature maps from each vehicle were compressed through quantization to reduce communication costs. This on-demand sharing mechanism enabled efficient cooperation among connected vehicles by selectively transmitting compressed feature maps; however, the integration of spatial-temporal information was not considered. In CPSC \cite{CP_based_on_ST_feature}, feature-level compression was performed by focusing on critical regions of perceptual information in the spatial-temporal domain, and the data transmission strategy was adapted based on network conditions.

Feature encoding using Convolutional Neural Networks (CNNs) has been widely adopted for feature compression in cooperative perception. In FS-COD \cite{marvasti2020cooperative}, a filter was applied at the final layer of the CNN-based feature extractor to control the size of the data shared between cooperative vehicles. In DiscoNet \cite{disconet2021li}, a $1\times1$ convolutional autoencoder \cite{masci2011stacked} was utilized to compress feature maps along the channel dimension, while in V2X-ViT \cite{v2xvit2022xu}, multiple $1\times1$ convolutional filters were applied for the same purpose. In COOPERNAUT \cite{cui2022coopernaut}, a Point Encoder based on the Point Transformer \cite{zhao2021point} was introduced to compress point clouds into compact representations, enabling the filtering of partially processed data to balance perception accuracy and communication load. Similarly, in CenterCoop \cite{CenterCoop}, informative cues from the full bird’s-eye view context were encoded into compact center representations, and communication costs were significantly reduced by transmitting these compact features as queries instead of the full feature map.

While CNN-generated feature maps offer potential data volume reduction, their dimensions often exceed transmission feasibility within current V2X technology scopes \cite{guo2022SlimFCP}. In \cite{guo2022SlimFCP}, the SENet channel attention module \cite{hu2018squeeze} was used to weigh channel importance, determining optimal channels for transmission based on attention weights and the uniqueness of semantic information. 
In EMIFF \cite{EMIFF}, a Feature Compression module was introduced, incorporating both channel and spatial compression blocks to enhance transmission efficiency. Although these approaches effectively reduced the required transmission bandwidth, the heterogeneity of transmitted information—caused by configuration disparities among agents—was not addressed \cite{FHM-ViT2023}.
To address this issue, a communication mechanism based on feature decoupling was introduced in What2comm \cite{yang2023what2comm}, in which bandwidth usage was strategically minimized by eliminating background noise to isolate sparse yet essential exclusive features, and by compressing common features along the channel dimension.

The aforementioned compression methods have uniformly applied the same compression ratio to all CAV data. However, this fairness approach fails to account for the varying contributions of CAVs to perception during data fusion \cite{magsino2022enhanced}. To address this limitation, SmartCooper \cite{zhang2024smartcooper} was proposed to optimize vehicle connectivity under communication constraints. In this method, a learnable encoder was used to dynamically adjust the compression ratio based on channel state information, and a judger mechanism was employed to filter out detrimental image data.

\subsection{V2X Communication}

\textcolor{black}{CP depends on V2X communication technologies to facilitate real-time information exchange among vehicles, infrastructure, and other road users. Dedicated Short-Range Communications (DSRC) and Cellular V2X (C-V2X) are two key technologies that support the low-latency and scalable communication needed in CP systems. Each technology has distinct features and challenges, which are explored below, followed by an analysis of how these limitations impact CP performance and the primary development bottlenecks in V2X communication. A comparison of the key technical differences between DSRC and C-V2X is presented in \cite{R1}.}

\subsubsection{\textcolor{black}{DSRC}}
\textcolor{black}{
DSRC was designed specifically for vehicle communication, adapting Wi-Fi technology into the IEEE 802.11p protocol \cite{REF802.11p}. Operating in the 5.9 GHz band, DSRC enables fast, two-way communication over short distances among vehicles, infrastructure, and other road users. Its decentralized design enables direct data exchange between vehicles without relying on a central network, minimizing latency \cite{REF802.11p}, but the communication range is limited to less than 1km, and the link capacity is less than 10 Mbps \cite{DSRC6}. Thus, DSRC can be used to implement Applications that require minimum latency, like collision warnings \cite{R1}.}

\textcolor{black}{Despite its early adoption, DSRC is increasingly being deprioritized in regulatory and industry roadmaps in favor of C-V2X, which offers greater scalability, network support, and integration with future 5G systems. 
Notably, the U.S. Federal Communications Commission (FCC) reallocated the 5.9 GHz spectrum in 2024, designating C-V2X as the primary technology for Intelligent Transportation Systems (ITS) and phasing out DSRC-based use in this band \cite{fcc2024}. 
}

\subsubsection{C-V2X}

C-V2X, standardized by the 3rd Generation Partnership Project (3GPP), enables broad V2X interactions encompassing V2V, V2I, V2N, and V2P connections. It operates in two modes defined in 3GPP Releases 14 through 16: Direct Communication (PC5 interface) and Network Communication (Uu interface) \cite{Release14, Release15, Release16}. The PC5 interface enables direct communication between vehicles and nearby infrastructure in the 5.9 GHz band, ideal for low-latency, localized applications such as crash avoidance and real-time safety alerts. The Uu interface, meanwhile, connects vehicles to broader V2N applications through LTE or 5G networks, supporting long-range data exchange for services like traffic management and infotainment, particularly for non-latency-sensitive applications.

\subsubsection{Impact of V2X on CP}

V2X communication enables data sharing, which is essential for CP, but it faces limitations that restrict CP’s full potential. Key challenges impacting CP performance are bandwidth and spectrum constraints, as well as latency and synchronization issues.

\paragraph{Bandwidth and Spectrum Constraints}

CP demands real-time, high-quality data, often involving high-resolution inputs from sensors such as LiDAR and radar, which places significant data rate requirements on communication systems. Achieving high data rates typically requires large bandwidth, stable transmission channels, and precise channel detection and estimation. DSRC operates within the 5.9 GHz frequency band with a total bandwidth of 75 MHz, divided into seven 10 MHz channels ($5.850 - 5.925$ GHz). However, due to high vehicle mobility, dynamic environmental changes, and multipath delay spread, DSRC’s communication channels are highly time- and frequency-selective \cite{DSRC5}. Studies estimate that up to 50\% of the coherence bandwidth is in the order of 1 MHz, with coherence times as short as 0.2 ms \cite{DSRC5}, limiting reliable transmission times and increasing dropout rates. Similar constraints affect C-V2X, which operates on a narrower 30 MHz spectrum ($5.895 - 5.925$ GHz) \cite{R3}. The PC5 mode of C-V2X offers an advantage by utilizing a blind hybrid automatic repeat request (HARQ) technique that combines turbo coding (providing better performance than the convolutional coding used in DSRC) with automatic repeat request error control and soft-combining \cite{R2}.

Addressing these challenges requires improved network planning and advanced channel detection and estimation techniques. For DSRC, an analytical model was proposed in \cite{DSRC4} to assess key metrics such as Channel Busy Ratio and Packet Delivery Ratio over distance, facilitating optimized network planning. 
Additionally, enhancements to IEEE 802.11p were made in \cite{DSRC1} through an OFDM-IM system in which pilot signals were strategically positioned and inactive channels were leveraged to increase detection accuracy. In \cite{DSRC2}, a low-complexity, high-speed deep neural network-augmented least square algorithm was introduced, resulting in reduced model size and latency, as well as improved accuracy under varied noise levels and channel conditions.
For C-V2X, a balanced resource allocation strategy was proposed in \cite{C-V2X-1}, in which resources were evenly distributed across subframes, leading to decreased packet loss and enhanced data throughput. Similarly, a pilot design incorporating zero-correlation zone sequences and a time-domain channel estimation approach was suggested in \cite{C-V2X-3} to mitigate co-channel interference in C-V2X systems.

\paragraph{\textcolor{black}{Latency and Synchronization}}

CP requires low latency and precise timing to ensure synchronized data fusion across vehicles. Although DSRC and C-V2X are both designed to minimize latency, high-density environments, network congestion, and physical obstructions can introduce delays. Studies indicate that DSRC and C-V2X can support applications with an end-to-end latency of approximately 100 milliseconds under moderate vehicular density \cite{R4}. However, CP demands stricter latency and high-quality service standards to function effectively. Additionally, synchronization challenges arise when fusing data from multiple agents, often leading to temporal mismatches that degrade CP accuracy. Both DSRC and C-V2X utilize the Global Navigation Satellite System (GNSS) for time synchronization, but DSRC operates with looser timing and asynchronous communication, while C-V2X adheres to stricter time standards \cite{R1}. Despite these efforts, network congestion can still impact latency and synchronization reliability.

To address congestion in DSRC, adjustments to vehicle settings have been proposed in \cite{DSRC3} based on information from supplementary channels, enabling lead vehicles to monitor hidden nodes within a group. Furthermore, a congestion-aware forwarding scheme was presented in \cite{DSRC7} to optimize multi-hop forwarding reliability and reduce latency under heavy network load. In C-V2X, a rate control mechanism has been introduced in \cite{C-V2X-3}, allowing power control to actively contribute to congestion management, while a DQN-based adaptive congestion control algorithm was suggested in \cite{C-V2X-4}, in which a central controller is used to train and distribute the policy, thereby improving QoS compliance over ETSI’s Distributed Congestion Control in dense traffic scenarios. For synchronization, a decentralized clock synchronization protocol was introduced in \cite{DSRC8} to enhance DSRC’s resilience against GPS spoofing and to improve reliability in the absence of GNSS signals. Meanwhile, a synchronization signal design using generalized reciprocal sequences for mmWave C-V2X was proposed in \cite{C-V2X-5}. This approach is shown to enhance OFDM-based V2V synchronization by achieving high detection accuracy, minimizing interference, and reducing computational complexity.

\subsection{Cooperative Information Alignment}
\textcolor{black}{
Cooperative information alignment is crucial for CP systems in autonomous driving to ensure that shared data from different sources is accurately represented within a common coordinate system \cite{allig2019alignment}. This process is essential for fusing information from multiple agents to achieve a unified understanding of the environment.}

\textcolor{black}{
\subsubsection{Data Alignment}
Each collaborative agent typically operates within its local coordinate system, which may differ in both position and orientation relative to the ego agent. Data shared from cooperative agents, whether raw data, intermediate features, or perception outcomes, is initially referenced in each agent's local frame.
To align the cooperative data from a collaborative agent (Agent B) with the ego vehicle (Agent A), the messages transmitted must include the sender's position and its orientation. Upon receiving data and pose information from the collaborative agent, the ego agent calculates a relative transformation matrix, \(T_\mathrm{B\rightarrow A}\), to map the incoming data into the ego agent's coordinate frame.}

\textcolor{black}{
To compute \(\mathbf{T}_\mathrm{B\rightarrow A}\), it is essential first to establish the transformation matrix \(\mathbf{T}\) that projects each agent's local coordinate system onto the global coordinate frame. This matrix incorporates both rotation and translation components, determined by each agent's global position \([x,y,z]\) and orientation angles \(\alpha\), \(\beta\), and \(\gamma\) (representing yaw, pitch, and roll, respectively). The transformation matrix \(\mathbf{T}\) is constructed as follows:
\begin{equation} \label{tmax}
\mathbf{T} =\begin{bmatrix}
  r_{11} &r_{12} &r_{13}  &x \\
  r_{21}&r_{22}  &r_{23}  &y \\
  r_{31}&r_{32}  &r_{33}  &z \\
  0&  0&  0&1
\end{bmatrix},    
\end{equation}
where \(r_{ij}\) (\(i, j \in \{1, 2, 3\}\)) denotes the entries of the 3D rotation matrix \(\mathbf{R}\), which can be calculated as:
\begin{equation} 
\boldsymbol{R}=\boldsymbol{R}_z(\alpha ) \cdot \boldsymbol{R}_y(\beta )\cdot \boldsymbol{R}_x(\gamma), 
\end{equation}
where \(\boldsymbol{R}_z(\alpha )\), \(\boldsymbol{R}_y(\beta )\), and \(\boldsymbol{R}_x(\gamma)\) are the basis rotation matrices around the \(z\), \(y\), and \(x\) axes, respectively:
\begin{equation}
\boldsymbol{R}_z(\alpha )=\begin{bmatrix}
  \cos(\alpha)&-\sin(\alpha)&0 \\
  \sin(\alpha)&\cos(\alpha)&0 \\
  0&0  &1
\end{bmatrix},
\end{equation}
\begin{equation}
\boldsymbol{R}_y(\beta )=\begin{bmatrix}
  \cos(\beta)&0&\sin(\beta) \\
  0&1&0 \\
  -\sin(\beta)&0&\cos(\beta)
\end{bmatrix},  
\end{equation}
\begin{equation}
\boldsymbol{R}_x(\gamma )=\begin{bmatrix}
  1&0&0 \\
  0&\cos(\gamma)&-\sin(\gamma) \\
  0&\sin(\gamma)&\cos(\gamma)
\end{bmatrix}.
\end{equation}
Based on (\ref{tmax}), the transformation matrices \(\mathbf{T}_\mathrm{A}\) and \(\mathbf{T}_\mathrm{B}\) for agents A and B can be derived from their respective poses. To find the relative transformation matrix \(\mathbf{T}_{\mathrm{B}\rightarrow\mathrm{A}}\), multiply the inverse of \(\mathbf{T}_\mathrm{B}\) by \(\mathbf{T}_\mathrm{A}\):
\begin{equation}
\mathbf{T}_{\mathrm{B}\rightarrow\mathrm{A}}=\mathbf{T}_\mathrm{B}^{-1}\cdot \mathbf{T}_\mathrm{A}.
\end{equation}
Each data point from agent B's perspective is represented as \(\mathbf{p}_\mathrm{B}=[a,b,c]\). Converting these 3D points to homogeneous coordinates by appending a \(1\) as the fourth element yields \(\mathbf{p}_\mathrm{B}=[a,b,c,1]^{\mathrm{T}}\). To align each data point with the ego vehicle's frame, multiply it by \(\mathbf{T}_{\mathrm{B}\rightarrow\mathrm{A}}\):
\begin{equation}
\mathbf{p} =\mathbf{T}_{\mathrm{B}\rightarrow\mathrm{A}} \cdot \mathbf{p}_\mathrm{B},
\end{equation}
where \(\mathbf{p}\) is the transformed point in the ego agent's coordinate frame. This alignment enables the direct fusion of incoming data from other agents with the ego's data, as all points are now within the same coordinate system.}

\subsubsection{Alignment for Pose Errors}
\textcolor{black}{
Accurate relative transformation matrices are foundational for ensuring alignment in a shared coordinate system. However, these matrices depend on sensor measurements from posture detection systems such as Global Positioning System (GPS) and Inertial Measurement Unit (IMU), which are prone to noise and accuracy limitations, potentially leading to misalignment and reduced perception performance \cite{su20233d}.}

\begin{table*}[!t]
\renewcommand{\arraystretch}{1.3}
\scriptsize
\caption{Summary of the Works of Cooperative Information Alignment for Pose Errors.}\label{tab_b}
\centering
\renewcommand\tabcolsep{3.8pt}
\begin{threeparttable}
\begin{tabular}{lccccccc}
\toprule
Method &Year &Scenario &Modality &Dataset/Simulator &Cooperation Style &Downstream Task &Available Code\\
\midrule
RANSAC-based \cite{leveraging2022yuan} &2022 &V2V &LiDAR &COMAP &Late &Detection, Location &$\times$\\
V2VNet$_{\mathrm{robust}}$ \cite{RobustV2VNet} &2021 &V2V &LiDAR &V2V-Sim &Intermediate &Detection, Motion Prediction &$\times$\\
CoAlign \cite{robust2022lu} &2022 &V2X &LiDAR &OPV2V,V2X-Sim,DAIR-V2X  &Intermediate &Detection &\href{https://github.com/yifanlu0227/CoAlign}{$\surd$} \\ 
\textcolor{black}{RoCo} \cite{RoCo} &2024 &V2X &LiDAR &DAIR-V2X, V2XSet &Intermediate &Detection &\href{https://github.com/HuangZhe885/RoCo}{$\surd$}  \\
ICP\&OP \cite{song2022efficient} &2022 &V2V &LiDAR &OPV2V &Late &Detection &$\times$\\
\textcolor{black}{V2X-PC} \cite{V2X-PC} &2024 &V2X &LiDAR &DAIR-V2X, V2XSet &Early &Detection &$\times$ \\
\textcolor{black}{CoBEVGlue} \cite{ni2024self} &2024 &V2X &LiDAR &OPV2V,DAIR-V2X,V2V4Real &Intermediate &Detection &\href{https://github.com/VincentNi0107/CoBEVGlue}{$\surd$} \\
Map Container \cite{jiang2022map} &2022 &V2X &Camera,LiDAR &VISSIM &Late &State Estimation &$\times$\\
\textcolor{black}{FreeAlign} \cite{lei2024robust} &2024 &V2X &LiDAR &OPV2V, DAIR-V2X &Intermediate &Detection &\href{https://github.com/MediaBrain-SJTU/FreeAlign}{$\surd$} \\
SCOPE \cite{yang2023spatio} &2023 &V2X &LiDAR &OPV2V,V2XSet,DAIR-V2X &Intermediate &Detection &\href{https://github.com/starfdu1418/SCOPE}{$\surd$} \\
V2X-ViT \cite{v2xvit2022xu} &2022 &V2X &LiDAR &V2XSet &Intermediate &Detection &\href{https://github.com/DerrickXuNu/v2x-vit}{$\surd$}\\
\textcolor{black}{EMIFF} \cite{EMIFF} &2024 &V2X &Camera &DAIR-V2X &Intermediate &Detection &$\times$  \\
CBM \cite{CBM2023} &2024 &V2V &Camera,LiDAR &SIND, OPV2V &Late &Detection &\href{https://github.com/zhyingS/CBM}{$\surd$} \\
\textcolor{black}{KM-based} \cite{SeeThrough} &2024 &V2V &Camera &KITTI, PreScan &Late &Detection &$\times$  \\
DMGM \cite{Gao2023DeepMG} &2023 &V2V &Camera &CARLA \& SUMO &Early &$-$ &\href{https://github.com/gaopeng5/DMGM}{$\surd$} \\
\bottomrule
\end{tabular}
\end{threeparttable}
\end{table*}

Many efforts have been undertaken to address pose errors, as summarized in Table \ref{tab_b}. A widely adopted approach involves statistical learning models. For instance, the method presented in \cite{leveraging2022yuan} corrected relative locations between two CAVs by matching key points in their perception results with the Random Sample Consensus (RANSAC) algorithm \cite{fischler1981random}. While effective, this approach required the setting of problem-specific thresholds. Based on V2VNet \cite{v2vnet2020wang}, a consistency module \cite{RobustV2VNet} was introduced to enhance relative pose estimations by structuring the global coherence of absolute poses as a Markov random field and incorporating Bayesian reweighting. Unfortunately, this model relies on ground-truth poses during the training phase, which may be limited in practical scenarios. 
In CoAlign \cite{robust2022lu}, an Agent-Object Pose Graph Optimization method was proposed to improve pose consistency through pose graph construction and optimization techniques. Similarly, in RoCo \cite{RoCo}, the pose correction problem was modeled as an object-matching task, wherein common objects detected by different agents were reliably associated, and agent poses were iteratively adjusted to minimize alignment errors. Both CoAlign and RoCo effectively mitigated relative pose errors without requiring precise supervision during the training process.

Point cloud registration methods typically focus on refining the Iterative Closest Point (ICP) algorithm \cite{besl1992method}, widely used for aligning 3D point clouds. In \cite{song2022efficient}, an ICP-based matching algorithm aligned two point sets, one representing 3D bounding boxes detected by the Ego and the other by the sender, with the relative transformation formulated as an optimal transport problem \cite{villani2009optimal} to minimize transportation costs between source and target points. 
In V2X-PC \cite{V2X-PC}, point clusters from spatial and temporal dimensions were aligned, and parameter-free solutions were proposed to adapt to varying noise levels without requiring finetuning. In CoBEVGlue \cite{ni2024self}, a spatial alignment module named BEVGlue was introduced to estimate relative poses between agents by matching co-visible objects without relying on an external localization system. In contrast to traditional point cloud registration methods, BEVGlue required only the transmission of bounding boxes and tracking identifiers, thereby reducing communication bandwidth while maintaining high-quality matching through verification of spatial relationships between matched nodes.

High-definition maps represent a valuable resource for achieving precise spatial alignment, primarily due to their ability to provide accurate self-vehicle localization. In this regard, a map-based CP framework called the ``map container" was introduced in \cite{jiang2022map}, which establishes benchmarks for spatiotemporal transformations by automatically mapping multi-agent perceptual information into the map coordinate space using a map-matching algorithm. 
However, the implementation of this algorithm was dependent on monocular cameras and GNSS/IMU devices.
In contrast, FreeAlign \cite{lei2024robust} was proposed to estimate relative poses by leveraging the invariant geometric structure formed by commonly perceived and observed objects across agents, thereby avoiding reliance on expensive high-end global localization systems.

Multi-scale feature interaction has been shown to effectively mitigate feature map misalignment caused by pose errors by encoding features at multiple scales and subsequently aggregating them into a final collaborative representation.
In this context, a spatiotemporal awareness CP framework, SCOPE \cite{yang2023spatio}, was designed to generate robust representations while accounting for localization errors through confidence-aware multi-scale feature interaction.  
Similarly, in V2X-ViT \cite{v2xvit2022xu}, a Multi-scale Window Attention Module was proposed to capture long-range spatial interactions at multiple scales through a pyramid of resolution windows, enabling both local and global spatial feature interactions and improving detection robustness against positioning errors. In EMIFF \cite{EMIFF}, the Multi-scale Cross Attention and Camera-aware Channel Masking modules were developed to counter the negative effects of pose errors by selectively integrating features, applying spatial offsets, and reweighting features based on camera parameters.

Inter-agent object association is aimed at mitigating localization errors in pose transformation by matching key points or objects from connected agents and determining the optimal relative pose transformation. To facilitate this, the Context-based Matching (CBM) algorithm \cite{CBM2023} was developed, in which object clusters were generated based on bounding box headings and correspondences were established through iterative pairwise comparison to maximize global consensus. Spatial differences were addressed in \cite{SeeThrough} using a data association and spatial error calibration method based on the Kuhn-Munkres algorithm. However, the identification of correspondences remains challenging due to visual folding, non-visible objects, and noisy perception. To overcome these issues, correspondence identification was framed as a graph-matching problem in \cite{Gao2023DeepMG}, and a masked GNN was employed to handle non-visible objects, explicitly accounting for occlusion and limited field-of-view challenges.

\subsection{Cooperative Information Fusion}
The effectiveness of CP crucially depends on the design of an efficient fusion strategy, which aims to consolidate shared information to provide coherent and accurate \textcolor{black}{perception results}. The following will examine recent fusion methods employed in CP information. For reference, Table \ref{tab_c} provides a summary of all discussed fusion models in this subsection.

\begin{table*}[!htp]
\renewcommand{\arraystretch}{1.3}
\scriptsize
\caption{Summary of the Works of Cooperative Information Fusion.}\label{tab_c}
\centering
\renewcommand\tabcolsep{3.8pt}
\begin{tabular}{cccccccc}
\toprule
Method &Year &Scenario &Modality &Dataset/Simulator &Cooperation Style &Downstream Task &Available Code\\
\midrule
Cooper \cite{cooper2019chen} &2019 &V2V &LiDAR &KITTI, T\&J &Early  &Detection &$\times$ \\
F-Cooper \cite{fcooper2019chen} &2019 &V2V &LiDAR &KITTI, T\&J &Intermediate &Detection &$\times$ \\
PillarGrid \cite{pillargrid2022bai} &2022 &V2X &LiDAR &CARTI &Intermediate &Detection &$\times$ \\
CoCa3D \cite{CoCa3D2023} &2023 &V2X &Camera &\makecell[c]{OPV2V$+$, DAIR-V2X,\\CoPerception-UAVs$+$} &Intermediate &Detection &\href{https://github.com/MediaBrain-SJTU/CoCa3D}{$\surd$} \\
\textcolor{black}{SiCP} \cite{SiCP} &2024 &V2X &LiDAR &OPV2V &Intermediate & Detection &$\times$ \\
\textcolor{black}{CoBEVFusion} \cite{qiao2023cobevfusion} &2023 &V2V &LiDAR \& Camera &OPV2V &Intermediate &Detection, Segmentation & $\times$ \\
\textcolor{black}{V2X-BGN} \cite{V2X-BGN} &2024 &V2X &Camera &V2XSet &Late &Detection &$\times$ \\
CPM-LCF \cite{mouawad2021collective} &2021 &V2X &LiDAR &$-$ &Late &Detection &$\times$\\
CI-based \cite{shan2022novel} &2022 &V2X &LiDAR &CARLA &Late &Tracking &$\times$ \\
\textcolor{black}{CPV-RCNN} \cite{teufel2023collective} &2023 &V2X &LiDAR &CARLA &Late &Detection &\href{https://github.com/ekut-es}{$\surd$} \\
\textcolor{black}{V2X-PC} \cite{V2X-PC} &2024 &V2X &LiDAR &DAIR-V2X, V2XSet &Early &Detection &$\times$ \\
V2VNet \cite{v2vnet2020wang} &2020 &V2V &LiDAR &V2V-Sim &Intermediate &Detection, Motion Prediction &$\surd$ \\
DiscoNet \cite{disconet2021li} &2021 &V2V &LiDAR  &V2X-Sim &Intermediate &Detection  &\href{https://github.com/ai4ce/DiscoNet}{$\surd$} \\
V2X-ViT \cite{v2xvit2022xu} &2022 &V2X &LiDAR &V2XSet &Intermediate &Detection &\href{https://github.com/DerrickXuNu/v2x-vit}{$\surd$} \\
HYDRO-3D \cite{HYDRO-3D} &2023 &V2X &LiDAR &V2XSet &Intermediate &Detection &$\times$ \\
\textcolor{black}{GraphAttention} \cite{GraphAttention} &2024 &V2X &LiDAR &V2X-Sim &Intermediate &Detection &$\times$ \\
CoBEVT \cite{xu2022cobevt} &2022 &V2V &Camera &OPV2V &Intermediate &Segmentation &\href{https://github.com/DerrickXuNu/CoBEVT}{$\surd$} \\
\textcolor{black}{QUEST} \cite{fan2023quest} &2024 &V2I & Camera &DAIR-V2X &Intermediate &Detection &$\times$ \\
COOPERANAUT \cite{cui2022coopernaut} &2022 &V2V &LiDAR &AUTOCASTSIM &Intermediate &Policy Learning &\href{https://github.com/UT-Austin-RPL/Coopernaut}{$\surd$} \\
\textcolor{black}{V2VFormer} \cite{V2VFormer} &2024 &V2V &LiDAR \& Camera &OPV2V, V2XSim, V2V4Real &Intermediate &Detection & $\times$ \\
\textcolor{black}{V2VFormer++} \cite{V2VFormer++} &2024 &V2V &LiDAR \& Camera &OPV2V, V2XSim &Intermediate &Detection & $\times$ \\
\textcolor{black}{MKD-Cooper} \cite{MKD_Cooper} &2023 &V2V &LiDAR &OPV2V, V2XSim, Experiment &Intermediate & Detection &\href{https://github.com/EricLee523/MKD-Cooper}{$\surd$} \\
\textcolor{black}{V2X-AHD} \cite{he2023v2x} &2023 &V2X &LiDAR &V2Xset &Intermediate & Detection &\href{https://github.com/feeling0414-lab/V2X-FKD}{$\surd$} \\
IoSI-CP \cite{liu2023rethinking} &2023 &V2X &LiDAR &OPV2V, V2XSet &Intermediate &Detection &\href{https://github.com/huangqzj/IoSI-CP/}{$\surd$} \\
Coop3D-Infra \cite{hybridfusion2020Arnold} &2022 &V2I &LiDAR &CARLA &Hybrid  &Detection &$\times$\\
ML-Cooper \cite{ML_Cooper} &2022 &V2V &LiDAR &KITTI &Hybrid &Detection &$\times$ \\
\textcolor{black}{Hybrid CP} \cite{Hybrid_CP} &2023 &V2V &LiDAR &OPV2V &Hybrid &Detection & $\times$ \\
\bottomrule
\end{tabular}
\end{table*}

\subsubsection{Traditional Fusion}

Primary reduction operators, such as summation, maximum, and average pooling, are widely used in cooperative information fusion for multi-agent perception. For example, the Cooper \cite{cooper2019chen} was developed for CAVs to enhance 3D object detection using raw point cloud data, though it involved high data overhead. 
This was extended in F-Cooper \cite{fcooper2019chen}, where feature-level fusion was employed to balance bandwidth constraints and detection accuracy through the use of max pooling for vehicle-to-vehicle voxel feature fusion.
Similarly, grid-wise max pooling was applied in PillarGrid \cite{pillargrid2022bai} to integrate deep features for cooperative three-dimensional object detection. In CoCa3D \cite{CoCa3D2023}, point-wise maximum fusion was utilized to combine bird’s-eye view (BEV) features from multiple CAVs, thereby enhancing camera-based detection via cooperative depth estimation.
In addition to these commonly adopted reduction operators, dedicated fusion modules have also been designed for fusing BEV features across CAVs. In SiCP \cite{SiCP}, a Dual-Perception Network (DP-Net) was proposed to assign different weights to LiDAR point clouds from different CAVs by extracting crucial gradient information, requiring only a small number of parameters. This concept was extended in CoBEVFusion \cite{qiao2023cobevfusion}, where a dedicated fusion module was employed to integrate heterogeneous features collected from the camera and LiDAR sensors on two CAVs.

As the late fusion strategy requires a minimum volume of exchanged information between CAVs, several traditional fusion approaches employing the late fusion strategy have been explored in practice. In the context of object detection, non-maximum suppression (NMS) serves as a non-parametric method for filtering overlapping bounding boxes based on confidence scores without requiring training or fitting parameters. 
In V2X-BGN \cite{V2X-BGN}, NMS was applied to aggregate redundant detection results from intelligent agents globally, based on 2D IoU in BEV space. 
A rule-based fusion scheme for cooperative perception messages (CPMs) was proposed in \cite{mouawad2021collective} to associate locally detected objects with those identified via vehicle-to-everything communication. This scheme was designed for scenarios in which CAVs generate CPMs according to the European Telecommunications Standards Institute (ETSI) specifications and employ a fusion centre to merge and broadcast the CPMs to nearby CAVs for subsequent perception tasks.

The non-parametric fusion methods offer minimal computational costs but often fail to capture data correlations. In contrast, parametric methods generally provide more accurate and reliable fusion by incorporating measurements and their uncertainties. For instance, the Covariance Intersection (CI) algorithm can consistently fuse estimates without prior knowledge of cross-correlations. 
Building on this concept, a CI-based fusion strategy for cooperative perception was introduced in \cite{shan2022novel}, in which local sensor data and object information received from other agents via V2X communication were consistently fused.

\textcolor{black}{In contrast to object-level and BEV-based approaches that often lead to feature degradation, the Point Cluster Aggregation (PCA) module in V2X-PC \cite{V2X-PC} preserves the low-level structural details of point clusters, thereby improving the accuracy of proposal bounding boxes. PCA integrates point clusters from multiple agents by matching them based on the distances between cluster centres and merging clusters that represent the same object into a new cluster. Its computational complexity is determined solely by the number of point clusters, without requiring convolution or padding operations, which makes it both effective and efficient for long-range collaborative perception.}

\subsubsection{Graph-based Fusion}
Multi-agent collaboration can be represented as a collaboration graph, with nodes as agent states and edges as pairwise collaborations. At the core is the GNN, which extracts relational data and generates node embeddings using local attributes and neighbouring information. Node states are continuously updated through inter-node communication and aggregation.

In V2VNet \cite{v2vnet2020wang}, each vehicle used a fully connected GNN \cite{schlichtkrull2018modeling} as an aggregation module for data fusion from various viewpoints. Messages were aggregated at each node using a mask-aware accumulation operator, considering overlapping Field of Views (FoVs). However, it employed a scalar-valued edge weight to indicate the importance of the neural message from one agent to another, which could not effectively model the strength of collaboration between two agents. 
In contrast, in DiscoNet \cite{disconet2021li}, an edge encoder was employed to establish correlations between feature maps from different agents by generating matrix-valued edge weights. Feature maps were concatenated and processed through convolutional layers for downsampling, followed by a softmax operation for normalization. Additionally, knowledge distillation was used to enhance training by aligning the intermediate features of a fusion-based student model with those of an early fusion-based teacher model, thereby improving feature abstraction and object detection performance.

The Heterogeneous Multi-agent Self-attention (HMSA) module in V2X-ViT \cite{v2xvit2022xu} was designed to fuse cooperative features by associating agent types with nodes and edges in a directed graph, and by computing importance weights between nodes accordingly. To further enhance cooperative perception performance, HYDRO-3D \cite{HYDRO-3D} was developed as an extension of V2X-ViT to construct a hybrid object detection and tracking model. In this framework, detection features from V2X-ViT were combined with historical tracking data extracted using a spatial-temporal pyramidal three-dimensional network. The fused features were shown to improve robustness and enhance detection accuracy, as demonstrated through experimental evaluation.

Traditional feature fusion methods, such as concatenation and summation, offer only fixed linear fusion and do not account for the varying importance of feature maps. To address this limitation, a graph attention feature fusion network was proposed in \cite{GraphAttention}, in which important regions within feature maps were selectively emphasized through channel and spatial attention based aggregation. Through this approach, informative areas were adaptively highlighted, leading to enhanced feature representation for the ego vehicle and improved object detection precision.

\subsubsection{Attention-based Fusion}

The attention mechanism introduced in the transformer model has been widely adopted for determining fusion weights.
The Point Transformer \cite{zhao2021point} was shown to effectively learn compact representations from three-dimensional point clouds, demonstrating strong capability in capturing non-local interactions and producing permutation-invariant representations. This made it well-suited for aggregating LiDAR point clouds from multiple agents. 
In COOPERNAUT \cite{cui2022coopernaut}, a point transformer-based representation aggregator was introduced for information fusion, consisting of a voxel max-pooling operation for spatial aggregation and a point transformer block for fusing multi-agent perception data. 
Similarly, in V2VFormer \cite{V2VFormer} and V2VFormer++ \cite{V2VFormer++}, attentions based on channel and spatial information were employed to fuse features extracted from multiple CAVs. 
Together with knowledge distillation technology, several attention modules were employed to effectively transfer valuable knowledge from multiple excellent teacher models in different views and then aggregated into the student model on the ego CAV \cite{he2023v2x, MKD_Cooper}.

Given the high cost of LiDAR sensors, the study in \cite{xu2022cobevt} proposed a sparse vision transformer framework, CoBEVT, which uses multi-agent multi-camera features for collaborative BEV segmentation. CoBEVT introduced a 3D vision transformer that fuses BEV features from multiple agents using the Fused Axial attention module, which integrates global and local attention mechanisms to capture both long-range and detailed regional interactions. QUEST \cite{fan2023quest} followed the variable query concept first proposed by DETR \cite{DETR} and extended it to V2I CP settings for camera-based 3D object detection.

The aforementioned attention-based models have significantly advanced cooperative information fusion by focusing primarily on spatial dimensions. However, their reliance on single-frame prediction neglects historical context and overlooks important temporal semantic cues, limiting their ability to detect fast-moving objects due to point cloud sparsity. 
To address these limitations, a Context-aware Information Aggregation model was introduced in SCOPE \cite{yang2023spatio}, in which a pyramid LSTM network was employed to incorporate spatiotemporal data from previous frames of the ego agent, generating refined context-aware features.
These features were fused with cooperative features based on complementary contributions and spatial attention maps obtained through max pooling, which may not fully capture correlations among informative features. 
To overcome this shortcoming, the Historical Prior Hybrid Attention (HPHA) fusion algorithm was proposed in IoSP-CP \cite{liu2023rethinking}, comprising a multi-scale transformer and a short-term attention module. In this framework, attention weights were extracted at different spatial scales using the multi-scale transformer, while the short-term attention module was used to capture temporal dependencies and identify discriminative features, thereby enhancing informative features and suppressing background noise.

\begin{table*}[!t]
\renewcommand{\arraystretch}{1.3}
\scriptsize
\caption{Summary of the Works on Issues of Cooperative Perception and Corresponding Methods.}\label{tab_5}
\centering
\renewcommand\tabcolsep{4pt}
\begin{threeparttable}
\begin{tabular}{clccccccc}
\toprule
Category &Method &Year &Scenario &Modality &Dataset/Simulator &Cooperation style &Downstream task &Available Code\\
\midrule
\multirow{4}{*}{MH} &MAMP \cite{chen2022model} &2022 &V2V &LiDAR &OPV2V &Late &Detection &\href{https://github.com/DerrickXuNu/model_anostic}{$\surd$} \\ 
&UDT2T \cite{allig2020unequal} &2022 &V2V &$-$ &CARLA &Late &Tracking &$\times$\\
&MPDA \cite{xu2022bridging} &2022 &V2X &LiDAR &V2XSet &Intermediate &Detection  &\href{https://github.com/DerrickXuNu/MPDA}{$\surd$} \\ 
&\textcolor{black}{MACP} \cite{MACP} &2024 &V2V  &LiDAR  &V2V4Real, OPV2V  &Intermediate  &Detection &\href{https://github.com/PurdueDigitalTwin/MACP}{$\surd$}  \\ 
\cmidrule{1-9}
\multirow{5}{*}{DH} 
&HM-ViT \cite{FHM-ViT2023} &2023 &V2V &Camera\&LiDAR &OPV2V &Intermediate &Detection &\href{https://github.com/XHwind/HM-ViT}{$\surd$}\\
& \textcolor{black}{DI-V2X} \cite{Domain_Invariant_CP_3DOD} &2024 &V2X &LiDAR &DAIR-V2X, V2XSet &Intermediate &Detection &\href{https://github.com/Serenos/DI-V2X}{$\surd$} \\
&\textcolor{black}{DG-CoPerception} \cite{DomainGeneralizationCP} &2023 &V2V &Camera &OPV2V &Intermediate &Segmentation &\href{https://github.com/DG-CAVs/DG-CoPerception}{$\surd$}\\
&VINet \cite{bai2022vinet} &2022 &V2X &LiDAR &CARTI &Intermediate &Detection &$\times$\\ 
&\textcolor{black}{V2X-M2C} \cite{V2X-M2C} &2024 &V2X &LiDAR &V2XSet, OPV2V &Intermediate &Detection &$\times$\\ 
&\textcolor{black}{VIMI} \cite{VIMI} &2023 &V2I & Camera &DAIR-V2X &Intermediate &Detection &\href{https://github.com/Bosszhe/VIMI}{$\surd$} \\
\cmidrule{1-9}
{MH \& DH} &\textcolor{black}{HEAL} \cite{lu2024extensible} &2024 &V2X &Camera\&LiDAR &OPV2V-H, DAIR-V2X &Intermediate &Detection &\href{https://github.com/yifanlu0227/HEAL}{$\surd$} \\
\cmidrule{1-9}
\multirow{4}{*}{LC} 
&V2VAM$+$LCRN \cite{li2022learning} &2022 &V2V &LiDAR &OPV2V &Intermediate &Detection &$\times$\\
&V2X-INCOP \cite{V2X-INCOP2023} &2024 &V2X &LiDAR &V2X-Sim,OPV2V,DAIR-V2X &Intermediate &Detection &$\times$\\
&\textcolor{black}{SsAW} \cite{CP_V2V_Communication_Self_Supervise} &2024 &V2V &LiDAR &OPV2V, V2V4Real &Intermediate &Detection &$\times$ \\ 
&\textcolor{black}{TempCoBEV} \cite{Unlocking_Past_Information} &2024 &V2V &Camera &OPV2V &Intermediate &Segmentation &$\times$\\
\cmidrule{1-9}
\multirow{8}{*}{CD} 
&\textcolor{black}{CooperFuse} \cite{CooperFuse} &2024 &V2X &LiDAR &V2X-real &Late &Detection &$\times$ \\
&FFNet \cite{Yu2023VehicleInfrastructureC3} &2023 &V2I &LiDAR &DAIR-V2X &Intermediate &Detection &\href{https://github.com/haibao-yu/FFNet-VIC3D}{$\surd$} \\
&SyncNet \cite{syncnet2022lei} &2022 &V2V &LiDAR &V2V-Sim &Intermediate &Detection &\href{https://github.com/MediaBrain-SJTU/SyncNet}{$\surd$} \\
&\textcolor{black}{CoBEVFlow} \cite{AsynchronyRobust} &2023 &V2V &LiDAR &CARLA, DAIR-V2X &Intermediate &Detection & {\href{https://github.com/MediaBrain-SJTU/CoBEVFlow}{$\surd$}} \\
&V2VNet \cite{v2vnet2020wang} &2020 &V2V &LiDAR &V2V-Sim &Intermediate &\makecell[c]{Detection,\\Motion Prediction} &$\times$ \\
&V2X-ViT \cite{v2xvit2022xu} &2022 &V2X &LiDAR &V2XSet &Intermediate &Detection &\href{https://github.com/DerrickXuNu/v2x-vit}{$\surd$}\\
&\textcolor{black}{V2X-PC} \cite{V2X-PC} &2024 &V2X &LiDAR &DAIR-V2X, V2XSet &Early &Detection &$\times$ \\
&\textcolor{black}{FreeAlign} \cite{lei2024robust} &2024 &V2X &LiDAR &OPV2V, DAIR-V2X &Intermediate &Detection &\href{https://github.com/MediaBrain-SJTU/FreeAlign}{$\surd$} \\
\cmidrule{1-9}
\multirow{1}{*}{LC \& CD} &\textcolor{black}{Ro-temd} \cite{RobustCPTemporalInformationDisturbance} &2024 &V2V &LiDAR &OPV2V &Intermediate &Detection &\href{https://github.com/hexunjie/Ro-temd}{$\surd$}\\
\cmidrule{1-9}
\multirow{6}{*}{DS} 
&\textcolor{black}{AdversAttack} \cite{tu2021adversarial} &2021 &V2V &LiDAR &V2V-Sim &Intermediate &Detection &$\times$\\
&\textcolor{black}{ROBOSAC} \cite{li2023among} &2023 &V2X &LiDAR &V2X-Sim &Intermediate &Detection &\href{https://github.com/coperception/ROBOSAC}{$\surd$} \\
&TrustEsti \cite{allig2019trustworthiness} &2019 &V2V &Camera/LiDAR &$-$ &$-$ &$-$ &$\times$\\
&FDII \cite{Zhang2023CooperativePF} &2023 &V2V &LiDAR &CARLA &Early &Detection &$\times$\\
&\textcolor{black}{CAD} \cite{zhang2023data} &2024 &V2V &LiDAR &Adv-OPV2V, Adv-MCity &Early/Intermediate &Detection &\href{https://github.com/zqzqz/AdvCollaborativePerception}{$\surd$}\\
&\textcolor{black}{MADE} \cite{zhao2023malicious} &2024 &V2V &LiDAR &V2X-sim, DAIR-V2X &Intermediate &Detection &$\times$\\
\cmidrule{1-9}
\multirow{4}{*}{DP}
&\textcolor{black}{FedAD} \cite{FedAD} &2023 &V2I &Camera &Udacity &$-$ &\makecell[c]{Turning\\Angle Prediction} &$\times$ \\
&FedBEVT \cite{FedBEVT2023} &2023 &V2X &Camera &CARLA\&OpenCDA &$-$ &Segmentation &\href{https://github.com/rruisong/FedBEVT}{$\surd$} \\
&FL-CCS \cite{song2023v2x} &2023 &V2X &$-$ &$-$ &$-$ &$-$ &$\times$ \\
&\textcolor{black}{FedDWA} \cite{FedDWA} &2024 &V2X &Camera &CARLA\&OpenCDA &$-$ &Segmentation &$\times$\\
\cmidrule{1-9}
\multirow{8}{*}{PU} &CoPEM \cite{piazzoni2022copem} &2022 &V2X &$-$ &ViSTA & Late &Detection &$\times$ \\
&DMMCP \cite{cai2023consensus} &2023 &V2X &Camera/LiDAR &$-$ &Late &Detection, Tracking &$\times$ \\
&\textcolor{black}{CMP} \cite{wu2024cmp} &2024 &V2V &LiDAR &OPV2V &Intermediate &Motion Prediction &$\times$ \\
&Double-M \cite{Su2023uncertainty} &2023 &V2X &LiDAR &V2X-Sim &\makecell[c]{Early\\/Intermediate/Late} &Detection &\href{https://github.com/coperception/double-m-quantification}{$\surd$} \\
&GevBEV \cite{Yuan2023GeneratingEB} &2023 &V2V &LiDAR &OPV2V &Late &Detection, Segmentation &$\times$ \\
&MOT-CUP \cite{su2023collaborative} &2024 &V2X &LiDAR &V2X-Sim &Late &Detection, Tracking &$\times$\\
&\textcolor{black}{DMSTrack} \cite{PointPillar+DMSTrack} &2024 &V2V &LiDAR &V2V4Real &Late &Tracking &\href{https://github.com/eddyhkchiu/DMSTrack}{$\surd$}\\
&\textcolor{black}{GCDA} \cite{GCDA} &2024 &V2V &LiDAR &V2V4Real &Late &Tracking &$\times$\\
\cmidrule{1-9}
\multirow{3}{*}{TD} &STAR \cite{li2022multirobot} &2022 &V2X &LiDAR &V2X-Sim &Intermediate &\makecell[c]{Scene completion,\\Detection,Segmentation}  &\href{https://github.com/coperception/star}{$\surd$} \\
&CORE \cite{wang2023core} &2023 &V2V &LiDAR &OPV2V &Intermediate &\makecell[c]{Scene completion,\\Detection,Segmentation} &\href{https://github.com/zllxot/CORE}{$\surd$} \\
\cmidrule{1-9}
\multirow{2}{*}{S2R} &S2R-ViT \cite{li2023s2r} &2024 &V2V &LiDAR &OPV2V,V2V4Real &Intermediate &Detection &$\times$ \\
&\textcolor{black}{DUSA} \cite{kong2023dusa} &2023 &V2X &LiDAR &V2XSet, DAIR-V2X &Intermediate &Detection &$\times$\\
\bottomrule
\end{tabular}
\begin{tablenotes} 
\scriptsize
\item[] \textbf{Note:} MH: Model Heterogeneity, DH: Data Heterogeneity, LC: Lossy Communication, CD: Communication Delays, DS: Data Security, DP: Data Privacy, PU: Perception Uncertainty, TD: Task Discrepancy, S2R: Simulation to Reality.
\end{tablenotes}
\end{threeparttable}
\end{table*}

\subsubsection{Hybrid Fusion}
Different collaboration schemes have advantages and limitations and can complement each other effectively. However, single-type fusion methods may not fully exploit the potential for \textcolor{black}{both deep fusion of various data types and consuming minimum communication bandwidth by generating higher-level collaborative representations}. Hybrid fusion-based approaches offer promise in harnessing the strengths of diverse collaboration schemes by integrating multiple fusion methods within a single CP model, \textcolor{black}{thus balancing the accuracy and transmission data size}.

Objects near the onboard sensor typically exhibit a high point density and are more easily detected using data from a single sensor, while distant points tend to be sparse due to limited sensor visibility. Early collaboration for distant regions can improve detection accuracy, as demonstrated by research \cite{hybridfusion2020Arnold}, which proposed a V2I cooperative 3D object detection model using hybrid fusion. This method transmits raw data (early fusion) for low-density areas and shares object-level information (late fusion) for nearby areas, providing a cost-effective solution but with limited flexibility in adjusting data sharing. 
In Hybrid CP \cite{Hybrid_CP}, a region-based hybrid collaborative perception strategy was proposed, in which collaboration was classified into two types based on the overlap between the detection ranges of vehicles. For overlapping areas, feature-level intermediate collaboration was performed by exchanging and fusing features from multiple CAVs. In contrast, late collaboration was employed in non-overlapping areas by generating and sharing local detection results with low communication volume. To enhance adaptability in dynamic vehicular networks, ML-Cooper \cite{ML_Cooper} was developed to optimize V2V bandwidth by dynamically adjusting data sharing across raw, feature, and object levels according to current channel conditions.

\section{Issues of Cooperative Perception and Corresponding Methods} \label{issue_of_cooperative_perception}

While the majority of research on CP for driving automation has focused on key aspects of the collaboration process, such as communication latency, information alignment, and fusion strategies, as discussed in the previous section, limited attention has been given to subtle yet crucial challenges that have practical implications for advancing Cooperative ITS. This section delves into exploring recent findings and the corresponding methodologies to address these practical challenges. Table \ref{tab_5} provides a comprehensive overview of characteristics related to these advancements.

\subsection{Agent Heterogeneity}
Although CP systems have made significant progress, it is important to acknowledge that these improvements often assume that the agents involved are uniform. In reality, it is difficult to achieve uniformity in agent configuration and sensor equipment across connected agents due to varying preferences among system developers and automobile manufacturers. 
Therefore, it is necessary to prioritize finding solutions to the practical challenges of agent heterogeneity, including model and data discrepancies among agents.

\subsubsection{Model Heterogeneity}

Sharing perception models with identical parameters among agents could raise concerns about privacy and confidentiality, especially when CAVs come from various automotive manufacturers. Even within a single automotive enterprise, variations in vehicle types and model update frequency can result in differences in perception among model versions. However, using different models for individual agents can create domain gaps in shared perception data. Neglecting model heterogeneity can significantly reduce the effectiveness of CP, thereby diminishing the benefits of multi-agent cooperation.

Object-level cooperative perception frameworks eliminate the need for full model data exchange by sharing perception outputs such as bounding boxes, confidence scores, and participating agents \cite{chen2022model}. However, due to model heterogeneity, the confidence scores generated by different agents may not always align accurately. To address this issue, a model-agnostic cooperative perception framework was proposed in \cite{chen2022model}, in which a Doubly Bounded Scaling calibrator was used to eliminate bias in the confidence scores, and a Promote-suppress Aggregation algorithm was employed to select high-confidence bounding boxes based on calibrated scores and spatial consistency.

In addition to variations in confidence levels, differences in motion prediction models may lead to discrepancies in detection outcomes among connected agents, particularly when some agents lack information such as acceleration or yaw rate.
To address heterogeneous perception fusion with varying state spaces, a CI-based track-to-track fusion approach was proposed in \cite{allig2020unequal}, augmenting low-dimensional tracks and exploring methods for state augmentation.

Late collaboration preserves model privacy by enabling agents to train models individually and share only their outputs. However, this approach may introduce noise and degrade performance.
Intermediate collaboration is more flexible and commonly adopted in V2X CP, but differences in feature extractors can result in a domain gap.
To address this issue, a Multi-agent Perception Domain Adaptation (MPDA) framework was introduced in \cite{xu2022bridging}, in which a feature resizer and a sparse cross-domain transformer were incorporated to align and produce domain-invariant features, thereby minimizing disparities between agents utilizing heterogeneous feature extraction networks. 
In \cite{MACP}, a Convolution Adapter (ConAda) was introduced to mitigate domain shifts by projecting the source feature map to a lower dimension and subsequently transforming it back. Furthermore, a single-agent pre-trained model was adapted to cooperative settings by freezing most parameters and integrating lightweight modules. Although this approach effectively mitigated domain shifts and enhanced perception, reduced precision for objects within the 50–100m range was observed, attributed to inherent biases in the pre-trained model.

\subsubsection{Data Heterogeneity}

In practical autonomous driving, connected agents often have different sensor modalities: LiDAR provides precise geometry, while cameras offer rich semantics. Collaborating with agents with diverse sensors enhances perception by leveraging complementary information. However, sharing data between camera and LiDAR agents introduces semantic disparities, and measurements differ between roadside infrastructure and CAVs. Additionally, sensor noise varies due to maintenance and hardware precision, posing challenges for developing a heterogeneous cooperative system.
 
In a multi-agent hetero-modal setting, connected agents have various sensors, forming a dynamic heterogeneous graph with stochastic sensor types and changing poses across scenes. 
To address these challenges, HM-ViT \cite{FHM-ViT2023} was proposed, in which the H3GAT module was introduced for vehicle-to-vehicle cooperative three-dimensional object detection. 
In this framework, node states were updated by extracting inter-agent and intra-agent cues using transformer blocks and a hetero-modal MLP layer. 
This design allowed camera and LiDAR features to be processed with distinct parameters, enabling greater flexibility, robustness, and improved performance in scenarios involving diverse sensor modalities.

On the other hand, identical modality sensors can vary in brand, resolution, and configuration, leading to domain discrepancies that impact system performance. 
In response to this challenge, DI-V2X \cite{Domain_Invariant_CP_3DOD} was proposed, in which a teacher-student distillation architecture was introduced, consisting of three key components: the Domain-Mixing Instance Augmentation module was designed to align data distributions between the teacher and student models, the Domain-Adaptive Fusion module was employed to combine vehicle and infrastructure features, and the Progressive Domain-Invariant Distillation module was used to align features from the student and teacher models both before and after fusion. Through this design, a generalized representation across various domains was achieved, thereby improving V2X perception performance.

Besides sensor characteristics, diverse environmental conditions inevitably introduce a substantial domain gap and data heterogeneity among CAVs \cite{DomainGeneralizationCP}. To address this, a domain generalization framework was proposed in \cite{DomainGeneralizationCP}, in which an amplitude augmentation module was employed to reduce the model's sensitivity to low-spectral variations, a domain alignment mechanism was used to unify image pixel distributions, and a meta-consistency training scheme was adopted to simulate domain shifts and optimize the model using consistency loss. This approach has the potential to eliminate domain discrepancies among CAVs prior to inference.

The distinctness in data characteristics across different agent types presents significant challenges in CP. 
This issue was addressed in VINet \cite{bai2022vinet}, where a Two-stream Fusion algorithm was employed to integrate heterogeneous features from infrastructures and CAVs through a feature mapping and max-out procedure performed on an edge server. 
In V2X-M2C \cite{V2X-M2C}, a heterogeneity-reflected convolution module was introduced to manage multiple heterogeneous agents using two parallel convolution pipelines, considering their different feature distributions while maintaining a lightweight and efficient architecture. However, time asynchrony and calibration errors between vehicle and infrastructure cameras can lead to inaccurate relative position detection, causing misalignment between ground truth and projected 2D bounding boxes. 
To address this, VIMI \cite{VIMI} was proposed, in which calibration noise was mitigated through the use of Multi-scale Cross Attention and Camera-aware Channel Masking modules. These modules applied multi-scale feature fusion and incorporated camera parameter priors to correct for misalignment caused by time asynchrony.

\subsubsection{Model and Data Heterogeneity}

While some existing studies address the heterogeneity of data and models, effectively modelling dynamically increasing heterogeneous scenarios continues to pose significant challenges. In reality, new heterogeneous agent types with different modalities and models may continuously emerge, causing domain gaps and limiting scalability. 
This issue was addressed in HEAL \cite{lu2024extensible}, where a unified feature space was first established for the initial homogeneous agents, and new heterogeneous agents were subsequently aligned to this space using a novel reverse alignment mechanism. Through this approach, the integration of new agent types was enabled with minimal training costs while preserving both model and data privacy.
Despite its scalability and efficiency, the reliance of HEAL on BEV features necessitated compatibility with keypoint-based LiDAR models, and its performance was shown to be dependent on accurate agent localization, which may be affected by localization noise in real-world scenarios.

\subsection{Lossy Communication}

In practical V2X communication, factors such as interference, tampering, Doppler shifts, routing failures, multipath effects, and signal fluctuations frequently cause incomplete or inaccurate information exchange, thereby diminishing the performance of CP.

Drawing inspiration from image-denoising networks, the Lossy Communication Aware Repair Network (LCRN) \cite{li2022learning} utilized an encoder-decoder architecture with skip connections to generate tensor-wise filters for repairing damaged features from lossy communication. These filters are applied to lossy features, which are then fused via a V2V attention module. While LCRN mitigates the impact of feature degradation, it does not directly address feature loss. 
In contrast, V2X-INCOP \cite{V2X-INCOP2023} was developed to recover missing data by leveraging historical information through an adaptive communication model that extracted multi-scale spatial-temporal features to estimate missing information, thereby countering communication interruptions.
Additionally, in \cite{RobustCPTemporalInformationDisturbance}, a Historical Frame Prediction (HFP) module was introduced to address information loss or delay issues by predicting the current frame with temporal associations, using the stacked results of the previous few historical frames as input.

Previous approaches often overlooked realistic channel models and failed to capture the complexity of real-world communication environments. 
To address this limitation, Rician fading with free-space path loss and the WINNER II model \cite{kyosti2007winner}, incorporating multi-path fading, were adopted in \cite{CP_V2V_Communication_Self_Supervise} to simulate practical communication channels. 
A self-supervised adaptive weighting model, based on a CNN with a Softmax layer, was introduced to mitigate channel distortion effects on intermediate fusion. The model was trained using contrastive self-supervision without manual annotations, and its generalization was validated across various data domains, detection backbones, noise levels, and path loss factors.

Access to historical information has been shown to reduce transmission frequency and provide a compensatory mechanism for potential communication failures, thereby enhancing system reliability and performance. Building on this concept, an importance-guided attention architecture was introduced in TempCoBEV \cite{Unlocking_Past_Information} to prioritize critical regions from the historical context, compensating for communication failures. This temporal module was designed to be integrated into state-of-the-art camera-based CP models without requiring complete model re-training, resulting in a significant reduction in training time. Experimental results on the OPV2V dataset demonstrated that TempCoBEV outperformed non-temporal models in predicting current and future BEV map segmentations, particularly in scenarios involving communication failures.

\subsection{Communication Delays}

Latency caused by network congestion has been shown to reduce perception accuracy and disrupt fusion alignment \cite{zhu2022latency}. A basic solution to mitigate latency involves adjusting timestamps by compensating for the time difference relative to a reference agent. Building on this concept, a multi-agent temporal synchronization module was introduced in CooperFuse \cite{CooperFuse}, in which time calibration communication, time compensation, and calibration completion were incorporated to minimize the effects of network transmission latency.

The time series prediction strategy has been demonstrated to be effective in addressing latency by estimating a sender’s perception at the current time step based on historical frames. In this context, FFNet \cite{Yu2023VehicleInfrastructureC3} was designed to leverage temporal coherence by extracting the first-order derivative of feature flow from previous infrastructure frames to predict future features and mitigate asynchrony. Although this first-order expansion method was effective over very short time intervals, it was found to have inherent limitations. In SyncNet \cite{syncnet2022lei}, a dual-branch pyramid LSTM network was employed to infer real-time features and cooperative attention. Similarly, the HFP module \cite{RobustCPTemporalInformationDisturbance} was developed to generate predictions of the current frame through a series of multidimensional convolutional mixing operations. However, these methods have not effectively addressed the challenges introduced by irregular time delays. As an alternative, each collaboration message was treated as an irregular sample in CoBEVFlow \cite{AsynchronyRobust}, and BEV flow map was estimated using historical frames for asynchronous feature alignment, although reliance on past data for latency compensation remained. In contrast, this dependency was bypassed in V2X-PC \cite{V2X-PC} by directly utilizing low-level coordinate information in point clusters to predict their positions at the current timestamp.

Re-encoding received features with delay time has been explored as an effective approach for mitigating communication delays. For instance, in V2X-ViT \cite{v2xvit2022xu}, time delay was addressed through a Delay-aware Positional Encoding module, in which sinusoidal functions conditioned on delay time and channel information were used to initialize a learnable embedding. Similarly, in V2VNet \cite{v2vnet2020wang}, a CNN was employed to manage latency by incorporating the received intermediate feature, relative 6DoF pose, and delay time to produce a time-delay-compensated representation. These methods assumed that latency was known. However, latency was typically computed using timestamps from different agents' clocks, which could introduce deviations and result in inaccurate measurements. To address this issue, a multi-anchor-based subgraph searching algorithm was introduced in FreeAlign \cite{lei2024robust} to identify a common subgraph between salient-object graphs across different agents, enabling estimation of clock deviations and determination of the precise time difference between collaborative messages. This approach ensured accurate temporal alignment without relying on external clock synchronization signals.

\subsection{Security and Privacy}
Data security and privacy are essential to the reliability of CP systems. V2X communication networks connecting various agents are vulnerable to potential attacks during data transmission, which could compromise system integrity. Furthermore, data sharing between agents introduces risks of privacy breaches, underscoring the need for robust security protocols within CP networks.

\subsubsection{Data Security}

While cooperative multi-agent systems offer significant promise, they also present security risks due to the potential for malicious or unreliable communication between agents \cite{Tsukada2022}. Malicious agents may tamper with shared packets or generate forged data. The ego vehicle is unable to independently verify missed events or confirm the authenticity of received data \cite{Tao2023zk-PoT}. 
Imperceptible perturbations introduced by adversaries can substantially alter perception outputs, compromising system integrity. 
Although adversarial training has been used to address these threats, it adds training overhead and often fails to generalize to novel attacks \cite{tu2021adversarial}. 
As a solution, ROBOSAC \cite{li2023among} was proposed, allowing ego vehicles to intelligently select trustworthy collaborators.
By detecting divergences in adversarial messages and aligning with reliable collaborators, ROBOSAC generalizes to unseen adversarial strategies and offers a flexible trade-off between performance and computational cost.

To ensure data security, assessing the quality of received information and detecting erroneous data before processing is crucial. Probabilistic modelling techniques have been utilized to estimate the trustworthiness of connected agents based on data consistency \cite{allig2019trustworthiness}. This approach applied a Bayes filter to quantify consistency, allowing the detection of forged information by analyzing the correlation between data consistency and agent reliability. 
However, this method was limited in detecting malicious agents, as it relied on substantial redundancy, typically requiring at least three conflicting sources, to identify attackers effectively.

FDII \cite{Zhang2023CooperativePF} was proposed to address spoofing attacks by leveraging LiDAR data from neighbouring vehicles to detect tampered points in a victim’s scan. Discrepancies were identified by comparing altered and unaltered scans, and attacks were classified using a decision tree algorithm, which was also used to update the unsafe areas associated with each vehicle. This approach was shown to be effective in identifying attack types and reconstructing vulnerable regions. However, decision trees were observed to struggle with complex data relationships and often exhibited bias toward the majority class in imbalanced datasets, potentially resulting in suboptimal classification of minority classes.

To counter data fabrication attacks, the Collaborative Anomaly Detection (CAD) model \cite{zhang2023data} was proposed, enabling benign vehicles to collectively detect malicious fabrications by sharing and merging occupancy maps. These maps, represented as fine-grained polygons, were shown to offer greater precision and flexibility compared to grid-based alternatives. Motion estimation was incorporated to track objects across frames, with motion data attached to the maps. Consistency checks were performed to identify discrepancies among synchronized maps and to compare individual perception results with the merged map, thereby raising alerts for conflicted regions indicative of potential anomalies. Although CAD was demonstrated to be effective in real-world scenarios, its efficacy was found to depend on the presence of at least one benign CAV monitoring the attacked region.

Malicious Agent Detection (MADE) \cite{zhao2023malicious} was developed to identify malicious agents using two key statistical metrics: match loss and collaborative reconstruction loss. Match loss was used to quantify discrepancies in bounding box proposals between the ego agent and the inspected agent, while collaborative reconstruction loss was employed to measure the consistency of feature maps when the two agents collaborated. These metrics were applied either independently or jointly to construct a multi-test with a controlled false detection rate. Although this approach provided valuable insights, its reliance on conformal p-values computed from calibration sets was based on the assumption that the ego agent was benign—an assumption that may not always hold in real-world scenarios.

\subsubsection{Data Privacy}
While collaboration among multiple agents shows great potential, it requires local data exchange between cooperating partners. In practice, connected agents often demonstrate heterogeneity and privacy concerns emerge as automotive manufacturers and software providers are generally unwilling to share proprietary data. This reluctance leads to incomplete and insufficient information for perception tasks, making it crucial to address privacy issues in V2X CP systems.

Federated Learning (FL)-assisted CAVs represent an emerging paradigm aimed at mitigating privacy concerns while reducing communication costs and diversifying the seen scenarios in comparison to conventional centralized learning \cite{9982368}. FL enables collaborating clients to effectively train models without the need for data aggregation or direct data exchange \cite{pandi2023federated}. However, determining the network resources and sensor placements for multi-stage training with multi-modal datasets in varied scenarios can be a complex challenge.

Conventional FL approaches have been challenged by data heterogeneity resulting from diverse sensor configurations. Variations in camera positions across clients have caused FL outcomes to deviate from the optimal values associated with local data. To address this issue, a federated transformer learning framework, FedBEVT \cite{FedBEVT2023}, was introduced, in which camera-attentive personalization was applied to alleviate the negative effects of data heterogeneity by privatizing the camera positional embedding. In \cite{FedDWA}, a Federated Dynamic Weighted Aggregation (FedDWA) algorithm, along with a dynamically adjusted loss function, was implemented to manage data heterogeneity in CP. However, network heterogeneity was not considered in these methods. Communication delays caused by varying connection qualities among clients have hindered the efficiency of the FL process. To overcome this limitation, a contextual client selection pipeline was proposed in \cite{song2023v2x}, where network conditions were modeled and FL communication latency was estimated under predictive transport scenarios using digital twin simulations. Through this approach, both data and network heterogeneity were effectively addressed, and communication efficiency was enhanced via context-aware client selection.

\textcolor{black}{
Challenges in FL for CAVs also include unfairness due to differences in data volumes and communication delays. To address the issues of imbalanced data distribution and variable channel conditions, the study in \cite{FedAD} employed a local training strategy customized for each CAV based on its data volume and channel conditions, thereby minimizing the number of global rounds needed and reducing the time intervals between rounds. Consequently, it promotes fairness in energy and time expenditure, accelerating model convergence while ensuring equitable costs for all CAVs.
}

\subsection{Perception Uncertainty}

Multi-agent cooperation markedly enhances the perception efficiency of autonomous vehicles. However, inherent uncertainties persist in CAVs due to out-of-range objects, sensor errors, and adverse weather conditions \cite{chawky2022evaluation}. These uncertainties can result in false detections, adversely affecting subsequent self-driving modules like planning and control. Addressing these uncertainties within CP is essential for ensuring safe autonomous driving.

Capturing perception system uncertainties is crucial for enhancing CAV safety. 
The Perception Error Model (PEM) \cite{piazzoni2021modeling} was developed to incorporate single-agent perception errors into the ground truth, thereby enabling the direct modeling of uncertainties within the simulation pipeline and eliminating the reliance on synthetic sensor data.
This approach was further extended in the Cooperative Perception Error Model (CoPEM) \cite{piazzoni2022copem}, which accounted for occlusion-related perception errors in vehicle-to-everything communication scenarios. CoPEM enabled performance tuning as the number of sensors increased. However, it required a statistical model for each agent and did not consider the high maneuverability of vehicle targets. 
To address this limitation, the Consensus-based Distributed Multi-Model Cooperative Perception (DMMCP) framework \cite{cai2023consensus} was proposed. 
It integrated prior knowledge from multiple motion models and adopted a hybrid consensus strategy using the Cubature Kalman Filter (CKF) \cite{jia2013high} to improve data fusion accuracy and manage model uncertainty. 
Despite its effectiveness, the initialization of the CKF posed challenges, particularly when initial state estimates were uncertain. 
More recently, CMP \cite{wu2024cmp} was introduced, which associated detected three-dimensional bounding boxes with trajectory segments. 
By aggregating cooperative perception and motion prediction data, CMP effectively mitigated propagated uncertainties and inaccuracies from upstream detection and tracking stages, resulting in enhanced motion prediction performance.

Uncertainty quantification is regarded as essential for ensuring safety in connected and autonomous vehicle (CAV) systems, as it allows the likelihood of outcomes to be estimated when aspects of the system remain uncertain. 
In Double-M \cite{Su2023uncertainty}, temporal features were utilized to address uncertainty in cooperative object detection. A customized Moving-Block Bootstrap algorithm was employed to model the multivariate Gaussian distribution for each bounding box corner, with auto-correlations in time-series data taken into account. Higher uncertainties were assigned to low-accuracy objects, resulting in improved detection accuracy. However, it was assumed that each bounding box corner followed an independent multivariate Gaussian distribution. A more comprehensive uncertainty modeling approach was proposed in GevBEV \cite{Yuan2023GeneratingEB}, in which a probabilistic BEV map was constructed using point-based spatial Gaussian distributions. Evidential Deep Learning \cite{sensoy2018evidential} was applied to estimate predictive probabilities and associated uncertainties, where the Gaussian densities were interpreted as evidence supporting a Dirichlet distribution. This allowed prediction uncertainty to be quantified and explained, and superior performance was demonstrated on the OPV2V benchmark.

In the context of object tracking, uncertainty quantification was addressed in the MOT-CUP framework \cite{su2023collaborative} by integrating detection uncertainty into the Kalman Filter (KF) and the object association process. Uncertainty was modeled using direct modeling techniques \cite{meyer2020learning} and conformal prediction methods \cite{angelopoulos2021gentle}, and subsequently applied to the Standard Deviation-based KF (SDKF) and a Negative Log-Likelihood (NLL) association refinement procedure. In SDKF, uncertainty was leveraged to enhance the accuracy of location prediction, while in NLL, tracking performance was improved by accounting for low-quality detections. Similarly, in DMSTrack \cite{PointPillar+DMSTrack}, a differentiable multi-sensor KF was utilized to estimate uncertainty for each detection, with a learned covariance model used to optimize tracking. However, multi-vehicle tracking information was not incorporated in either approach to enhance data association, and temporary tracking failures remained unresolved. To address this, a complementary data association module was introduced in \cite{GCDA} to identify and recover missed objects using information shared among CAVs. Experimental results demonstrated that the incorporation of uncertainty improved multi-object tracking performance.

\subsection{Task Discrepancy}

Most existing CP approaches are task-specific, necessitating retraining the entire model for different perception tasks. These task-specific methods have limitations and potentially hinder the models' generalization ability across a broader spectrum of perception tasks. The development of a CP framework independent of downstream tasks holds promise for advancing cooperative driving automation. In this context, recent endeavours have focused on a novel task known as multi-agent scene completion, also referred to as cooperative reconstruction. In this task, each agent learns to efficiently share information to reconstruct a complete scene observed by all agents. By training models to reconstruct holistic observations, this approach encourages the acquisition of more effective task-agnostic feature representations.

The Spatiotemporal Auto-Encoder (STAR) \cite{li2022multirobot} was designed to disregard downstream tasks and focus on reconstructing the entire scene through shared features in a self-supervised manner. In this framework, sub-sampled features are transmitted among agents to collectively cover the complete spatial region, allowing a balance to be maintained between scene reconstruction performance and communication costs. The reconstructed scene can subsequently be utilized for various downstream tasks without requiring additional training. This task-agnostic approach has been proposed as a novel direction in CP, enabling the decoupling of cooperation training from downstream task learning. However, due to incomplete scene reconstruction, a performance gap is still observed in downstream perception when compared to task-specific methods.

In contrast to STAR's self-supervised learning approach, the CORE method \cite{wang2023core}, learns to reconstruct comprehensive scenes under the guidance of BEV representations derived from the aggregation of raw sensor data from all agents. When provided with a group of agents and their 2D BEV maps, CORE conducts cooperative reconstruction through three main components: a compression module, a collaboration module, and a reconstruction module. The compression module calculates a compressed feature representation for each BEV, considering channel-wise compression and sub-sampling features along the spatial dimension. CORE incorporates a lightweight attention-aware collaboration module to facilitate information aggregation. The reconstruction module employs a decoder structure to regenerate a complete scene observation from the fused features. This ``learning-to-reconstruct" approach is task-agnostic and provides clear and meaningful supervision to encourage more effective cooperation, ultimately benefiting perception tasks.

\subsection{Simulation to Reality}
Due to the scarcity of real multi-agent data and the resource-intensive nature of manual labelling, most existing CP models often rely on simulated sensor data for training and testing. However, when these simulation-trained models are deployed in real-world settings, they frequently experience a decline in perception performance due to the substantial domain gap between simulation and reality. Domain adaptation techniques are employed to address these challenges by adapting models trained on labelled source domains to perform effectively in unlabeled target domains.

In contrast to the idealized simulation environment, real-world multi-agent scenarios introduce localization errors and communication latency among agents. Furthermore, the feature distribution in actual driving situations is notably more intricate than that in simulated data. To mitigate these two prominent disparities, the work in \cite{li2023s2r} introduced the Simulation-to-Reality (S2R) transfer learning framework, specifically S2R-ViT, which leverages labelled simulated data and unlabeled real-world data to minimize domain discrepancies in cooperative 3D object detection. The S2R-ViT framework comprises two key components: the S2R Uncertainty-aware Vision Transformer (S2R-UViT) and the S2R Agent-based Feature Adaptation (S2R-AFA). In S2R-UViT, two multi-head self-attention branches (LG-MSA) are utilized to incorporate both local and global attention mechanisms, enhancing feature interactions across spatial positions of all agents. Following LG-MSA, an uncertainty-aware module is introduced to assign weights to the ego-agent's features while considering uncertain factors stemming from the shared features of other agents. The S2R-AFA incorporates inter-agent and ego-agent discriminators, enabling S2R-ViT to generate robust, domain-invariant feature representations. Experimental results underscore the importance of domain adaptation for LiDAR-based 3D object detection when transitioning from simulation to real-world environments.

\textcolor{black}{
In addition to the domain gap between simulated and real-world data, there is also a domain gap among real-world collaborative agents, which further complicates the challenge of sim2real generalization. To tackle these issues, an unsupervised sim2real domain adaptation method named Decoupled Unsupervised Sim2Real Adaptation (DUSA) was proposed in \cite{kong2023dusa}. The DUSA solves the problems of sim2real adaptation and inter-agent adaptation by introducing the Location-adaptive Sim2Real Adapter (LSA) module and the Confidence-aware Inter-agent Adapter (CIA) module, respectively. The LSA module uses location importance to prompt the feature extractor to produce sim/real invariant features, while the CIA module uses confidence cues to help the feature extractor output agent-invariant features. This method effectively bridges the gap between simulated and real-world data for V2X collaborative perception, demonstrating its superiority over existing methods through experiments.}

\section{Issues for Dataset}
\label{dataset}

While CP has gained momentum in recent years, it remains a nascent field. One of the key obstacles impeding the progress of CP algorithms is the substantial cost and potential safety risks associated with conducting field experiments. Such experiments necessitate deploying numerous expensive CAVs and sizeable test areas. Additionally, acquiring precise annotations requires significant labour and time, especially when dealing with CP involving multiple agents. To address these challenges efficiently, an effective research strategy is to develop and validate CP algorithms within a simulated environment, which offers a cost-effective and safe alternative to real-world experimentation. However, validating V2X perception in authentic scenarios remains challenging due to the absence of well-established public benchmarks. This section delineates prominent simulation platforms and open-source datasets frequently employed in developing V2X CP systems.

\subsection{Simulation Tools}

Numerous simulation platforms have been developed to offer multi-modal sensory inputs and 3D annotations, thereby advancing the development of multi-agent CP before the widespread availability of real-world datasets. For example, Simulation of Urban MObility (SUMO) \cite{sumo2018} is a traditional traffic and driver behaviour simulator renowned for its capacity to handle large-scale and realistic traffic flows. It incorporates dynamic modelling for each vehicle, enabling users to swiftly create customized traffic scenarios through the Traffic Control Interface (TraCI) API. SUMO is highly convenient for integrating naturalistic trajectory data, which can be directly applied to the surrounding environment for testing CDA. However, it has limitations when providing high-resolution sensor data with exceptional fidelity.

In recent years, there has been a proliferation of autonomous driving simulators designed to facilitate high-fidelity modelling of the traffic environment and sensor capabilities. The following content will provide an overview of the fundamental characteristics of several prominent simulation tools used to develop and analyze CDA systems.

\subsubsection{CARLA}
CARLA \cite{carla2017} is a pioneering autonomous driving simulator celebrated for its versatility in specifying sensor suites and environmental conditions. This open-source platform was purpose-built to facilitate the development, training, and validation of novel autonomous driving systems. CARLA's architecture is notably scalable, employing a server-multi-client model to distribute computations across multiple nodes. The server continuously updates the environmental physics while users exert control over the client side through the CARLA API. Although CARLA permits environment extension, the process of swiftly creating scenarios and managing substantial traffic volumes has grown intricate. CARLA does offer a traffic manager module for generating background traffic. However, these modules rely on simplified behavioural rules that may not faithfully replicate actual driver behaviour.
 
\subsubsection{OpenCDA}
OpenCDA \cite{xu2023opencda} stands as an open and dedicated co-simulation framework designed for CDA. It encompasses both traffic and vehicles, offering a complete CDA vehicle software pipeline. OpenCDA seamlessly integrates CARLA and SUMO for lifelike scene rendering, vehicle modelling, and traffic simulation. This framework facilitates the evaluation of CDA within a CARLA + SUMO co-simulation environment and supports diverse levels and categories of cooperation among CAVs during simulation. OpenCDA also provides a comprehensive library of CDA research pipelines, source code, and standard autonomous driving components encompassing perception, localization, planning, control, and V2X communication modules. One of its key advantages is its flexibility, allowing users to easily substitute default modules with custom-designed ones to assess the impact of new functionalities on overall CDA performance.

\subsubsection{V2XP-ASG}
V2XP-ASG \cite{xiang2022v2xp} is designed to assess LiDAR-based V2X perception models by providing the capability to generate complex and realistic scenarios. This simulation framework comprises two key components: Adversarial Collaborator Search (ACS) and Adversarial Perturbation Search (APS). The ACS creates an adversarial collaboration graph based on an existing scene. It identifies and selects collaborators whose combined perspectives will likely lead to sub-optimal outcomes in perception tasks. The APS, on the other hand, focuses on perturbing the positions of vehicles within the scene. It achieves this using black-box optimization algorithms tailored to perception tasks' adversarial objectives. Combining these two stages in V2XP-ASG offers an efficient approach to enhance the complexity and difficulty of the scenarios presented in the simulation, thereby challenging the V2X perception models under evaluation.

\subsubsection{AUTOCASTSIM}
In \cite{cui2022coopernaut}, researchers introduced AUTOCASTSIM, an autonomous driving simulation framework built on the CARLA platform. AUTOCASTSIM was designed to evaluate cooperative driving models, particularly in accident-prone scenarios. This platform offers three benchmark traffic scenarios (Overtaking, Left Turn, and Red Light Violation) that impose decision-making challenges due to restricted line-of-sight sensing. AUTOCASTSIM includes an integrated networking simulation feature, facilitating customization of V2V communication dynamics. Additionally, it incorporates an advanced driving model that incorporates privileged environmental information, enabling the creation of customized and realistic traffic scenarios.

\subsubsection{CMM Co-simulation Platform}
The work in \cite{bai2023cyber} highlighted a common limitation in most current traffic simulators, which typically use intrinsic attributes of target objects without considering imperfect perception in CDA models. A CARLA-based co-simulation platform called the Cyber Mobility Mirror (CMM) was developed to address this limitation. The CMM platform enhances CDA by generating realistic perception data. A deep learning-based perception pipeline was integrated into the simulation system to establish a CMM framework capable of faithfully perceiving and reconstructing simulated traffic objects in 3D. This improvement enables CDA models to utilize perceived information from the CMM system, enhancing model fidelity and supporting more robust system validation, ultimately boosting overall confidence.

\begin{table*}[!t]
\renewcommand{\arraystretch}{1.3}
\caption{Comparison between Cooperative Perception-related Datasets.}\label{t_dataset}
\renewcommand\tabcolsep{3.5pt}
\centering
\begin{tabular}{cccccccccccccccc}
\toprule
{Category} &{Dataset} &{Year} &{Scenario} &\makecell[c]{Agent\\Number} &\makecell[c]{Annotation\\Range} &\makecell[c]{Object\\Category} &\makecell[c]{RGB\\Image} &\makecell[c]{LiDAR\\Frame} &\makecell[c]{3D\\Box} &\makecell[c]{Maps} &\makecell[c]{OD} &\makecell[c]{OT} &\makecell[c]{SS} &\makecell[c]{MP} &\makecell[c]{AP} \\
\midrule 
\multirow{9}{*}{Real-world} &DAIR-V2X-C \cite{dairv2x2022yu} &2022 &V2I &2 &280m &10 &39k &39k &464k &$\times$ &$\surd$ &$\times$ &$\times$ &$\times$ &$\times$ \\
&V2X-Seq \cite{v2x-seq} &2023 &V2I &2 &280m &10 &15k &15k &$-$ &$\surd$ &$\surd$ &$\surd$ &$\times$ &$\surd$ &$\times$ \\
&\textcolor{black}{HoloVIC} \cite{HoloVIC} &2024 &V2I &2 &70m &3 &100k &100k &11.5M &$\times$ &$\surd$ &$\surd$ &$\times$ &$\surd$  &$\times$\\
&V2V4Real \cite{Xu2023V2V4RealAR} &2023 &V2V &2 &200m &5 &40k &20k &240k &$\surd$ &$\surd$ &$\surd$ &$\times$ &$\times$ &$\times$ \\
&LUCOOP \cite{axmann2023lucoop} &2023 &V2V &3 &200m &5 &0 &54k &$-$ &$\surd$ &$\surd$ &$\surd$ &$\times$ &$\times$  &$\times$\\
&\textcolor{black}{TUMTrafV2X} \cite{TUMTrafV2X} &2024 &V2I &2 &200m &8 &5k &2k &30k &$\surd$ &$\surd$ &$\surd$ &$\times$ &$\surd$  &$\surd$\\
&\textcolor{black}{RCooper} \cite{Rcooper} &2024 &I2I &2 &230m &10 &50k &30k &$-$ &$\times$ &$\surd$ &$\surd$ &$\times$ &$\times$ &$\times$ \\
&\textcolor{black}{InScope} \cite{InScope} &2024 &I2I &2 &230m &4 &0 &21k &188k &$\times$ &$\surd$ &$\surd$ &$\times$ &$\times$ &$\times$\\
&\textcolor{black}{V2X-Real} \cite{V2X-Real} &2024 &V2X &4 &200m &10 &171k &33k &1.2M &$\surd$ &$\surd$ &$\times$ &$\times$ &$\times$ &$\times$ \\
\cmidrule{1-16}
\multirow{7}{*}{Similation} &CODD \cite{codd2022} &2022 &V2V &4--16 &100m &2 &0 &13.5k &204k &$\times$ &$\surd$ &$\surd$ &$\times$ &$\times$ &$\times$ \\
&OPV2V \cite{opv2v2022xu} &2022 &V2V &2--7 &120m &1 &44k &11k &233k &$\surd$ &$\surd$  &$\times$ &$\surd$ &$\times$ &$\times$ \\
&V2X-Sim \cite{v2xsim2022li} &2022 &V2X &2--5 &70m &1 &60k &10k &26k &$\surd$ &$\surd$ &$\surd$ &$\surd$ &$\times$ &$\times$\\
&V2XSet \cite{v2xvit2022xu} &2022 &V2X &2--7 &120m &1 &44k &11k &233k &$\surd$ &$\surd$ &$\times$ &$\times$ &$\times$ &$\times$ \\
&DOLPHINS \cite{mao2022dolphins} &2022 &V2X &3 &200m &2 &42k &42k &292k &$\times$ &$\surd$ &$\times$ &$\times$ &$\times$ &$\times$ \\
&DeepAccident \cite{DeepAccidentDataset2023} &2023 &V2X &5 &70m &6 &57k &57k &285k &$\times$ &$\surd$ &$\surd$ &$\surd$ &$\surd$ &$\surd$ \\
&\textcolor{black}{SCOPE} \cite{SCOPE} &2024 &V2X &3--24 &250m &5 &17.6k &17.6k &575k &$\times$ &$\surd$ &$\times$ &$\surd$ &$\times$ &$\times$ \\
\bottomrule
\end{tabular}
\begin{tablenotes}
\scriptsize  
\item[] \textbf{Note:} OD: Object Detection, OT: Object Tracking, SS: Semantic Segmentation, AP: Accident Prediction.
\end{tablenotes}
\end{table*}

\subsubsection{OpenCDA-ROS}
Existing CDA simulation platforms often exhibit disparities between simulation and real-world scenarios. To address this issue comprehensively, a unified tool and framework are needed to support CDA development in simulation and real-world environments. OpenCDA-ROS \cite{zheng2023opencda} has been developed as an extension of OpenCDA, tailored for the Robot Operating System (ROS). This adaptation serves as a bridge between simulated and real-world CDA, ensuring that algorithms and vehicles perform similarly in both settings. By reorganizing cooperative functions into a more shareable structure, OpenCDA-ROS benefits simulation and real-world prototyping and deployment, facilitating the seamless integration of these two domains in CDA development.


\subsection{Open Source Dataset}
The development of CP models relies on perception datasets gathered from various onboard sensors like radar, camera, and LiDAR. Utilizing these sensors for real-world perception tasks can be costly and time-consuming. A more economically viable alternative is to obtain cooperative data from simulation platforms featuring high-fidelity sensor modelling and perception capabilities. This subsection will outline several datasets relevant to CDA research, summarized in Table \ref{t_dataset}.

\subsubsection{Real-world Dataset}
A direct approach to evaluating V2X CP involves collecting diverse real-world test scenarios. DAIR-V2X-C \cite{dairv2x2022yu} is an autonomous driving dataset from real scenarios designed for research on V2I cooperation-related challenges. It comprises 38,845 LiDAR and 38,845 camera frames captured at intersections equipped with infrastructure sensors. Expert annotators have meticulously labelled all frames with 3D annotations. While DAIR-V2X supports the perception task, it does not cover the essential motion prediction aspect. V2X-Seq \cite{v2x-seq} offers a sequential V2I dataset, serving to develop and validate CP and motion forecasting algorithms. This dataset comprises two distinct sub-datasets: the sequential perception and trajectory forecasting datasets. The former includes over 15k frames from 95 scenarios, encompassing various data types such as infrastructure images, point clouds, vehicle-side images, point clouds, 3D detection/tracking annotations, and vector maps. The latter contains 80k infrastructure-view scenarios, 80k vehicle-view scenarios, and 50k cooperative-view scenarios captured from 28 intersections, with a substantial data coverage of 672 hours.

\textcolor{black}{DAIR-V2X-C and V2X-Seq employ a single viewpoint utilizing a camera and Lidar pair to gather data from various intersections. Nevertheless, the intricacy of traffic scenarios often results in frequent occlusion of targets captured by the camera, thus significantly restricting the efficacy of single-viewpoint roadside sensors. In contrast, HoloVIC \cite{HoloVIC} was developed through the establishment of multiple holographic intersections equipped with diverse sensor configurations, including cameras, lidars, and fisheye cameras. It comprises over 100k synchronized frames and provides annotations for 3D bounding boxes, object IDs, and global trajectories. Furthermore, it delineates five distinct tasks to foster advancements in related research areas.
}

The V2V4Real dataset \cite{Xu2023V2V4RealAR}, collected from two CAVs with multi-modal sensors in Columbus, Ohio, USA, covers diverse driving scenarios such as intersections, highways, and city streets. It includes 40k RGB frames, 20k LiDAR frames, 240k annotated 3D bounding boxes, and HDMaps for road topology prediction, supporting cooperative 3D object detection and Sim2Real adaptation. However, its two-vehicle setup limits its comprehensiveness for broader cooperative applications. Addressing this, LUCOOP \cite{axmann2023lucoop} provides data from three CAVs over a 4 km trajectory, including 54k LiDAR frames, 700k IMU measurements, 2.5 hours of GNSS data, UWB-based V2V and V2X measurements, and accurate ground truth poses. Although it offers a valuable resource for evaluating CP methods, it lacks complex traffic scenarios. \textcolor{black}{In contrast, the TUMTraf-V2X \cite{TUMTrafV2X} offers a more extensive collection, focusing on challenging traffic situations, varying weather conditions, and rare events, making it a more comprehensive tool for testing CP methods. It features 2k annotated point clouds, 5k images, and 30k 3D bounding boxes with track IDs.}

\textcolor{black}{
The datasets mentioned above primarily support either V2I or V2V cooperation, but not both modes together. In contrast, the V2X-Real dataset \cite{V2X-Real} is the first real-world dataset designed specifically for V2X CP research. It includes 33K LiDAR frames, 171K camera images, and over 1.2 million annotated 3D bounding boxes across ten object categories, collected from two CAVs and two smart infrastructures equipped with multi-modal sensors. The dataset is organized into four sub-datasets based on the ego agent type and collaboration mode: Vehicle-Centric, Infrastructure-Centric, V2V, and I2I CP. V2X-Real is collected in congested urban environments with dense traffic and pedestrian flow, providing rich and challenging scenarios for V2X CP research.}

\textcolor{black}{
All real-world datasets discussed so far have primarily focused on vehicle-centric cooperative tasks, such as V2V and V2I cooperation. However, roadside CP, particularly infrastructure-to-infrastructure (I2I) collaboration for comprehensive roadside coverage, remains unexplored. To address this, \cite{Rcooper} introduced RCooper, the first real-world dataset for roadside CP, featuring over 50k images and 30k point clouds with 3D annotations across various weather, lighting, and traffic conditions. InScope \cite{InScope} is another large-scale roadside CP dataset designed to tackle occlusion challenges in open traffic scenarios. It features multi-position LiDAR sensors deployed on infrastructure to broaden perception and complement vehicle-side systems, with 303 tracking trajectories and 187,787 3D bounding boxes captured over 20 days. InScope offers benchmarks for tasks like 3D object detection and multi-object tracking, providing comprehensive anti-occlusion algorithm evaluation. A new metric is also introduced to assess occlusion's impact on detection performance.}

\subsubsection{Simulation Dataset}
While realistic datasets allow CP models to generalize effectively to real-world driving situations, they are expensive, time-consuming, and limited in scenarios. A cost-effective alternative is to gather large-scale datasets in a high-fidelity simulator. 
The Cooperative Driving Dataset (CODD) \cite{codd2022}, generated using the CARLA simulator, comprises 108 snippets, each with 125 temporal frames of LiDAR point cloud data. Each frame contains LiDAR data from all vehicles in the scenario, sensor poses, and ground-truth annotations for 3D bounding boxes of cars and pedestrians. CODD focuses on a single driving scenario. In contrast, the OPV2V dataset \cite{opv2v2022xu} includes both camera images and LiDAR point clouds from connected vehicles across over 70 diverse scenes, spanning eight CARLA towns and Culver City, Los Angeles, using OpenCDA and CARLA. Both CODD and OPV2V provide ground truth for V2V cooperative 3D object detection, but neither includes roadside infrastructure, limiting their ability to evaluate V2I perception.

The V2X-Sim dataset \cite{v2xsim2022li} is a synthetic V2X-assisted CP dataset for autonomous driving, aimed at advancing research in multi-agent, multi-modality, and multi-task perception. It was generated using SUMO for realistic traffic flows and CARLA for capturing sensor streams from multiple vehicles at intersections. V2X-Sim provides synchronized recordings from roadside infrastructure and vehicles, along with comprehensive ground truth annotations for tasks such as detection, tracking, and segmentation. However, it lacks noise simulation and only includes a single type of road network. In contrast, the V2XSet dataset \cite{v2xvit2022xu} incorporates real-world noise during V2X collaboration using CARLA and OpenCDA, featuring over 11,447 frames divided into training, validation, and testing subsets. The DOLPHINS dataset \cite{mao2022dolphins}, generated with the CARLA simulator, encompasses six autonomous driving scenarios with V2X communication and temporally synchronized sensor data from an ego vehicle, a cooperating vehicle, and a roadside unit. It contains 42,376 frames of sensor data and 292,549 3D bounding box annotations for cars and pedestrians.

\textcolor{black}{
While the previously mentioned datasets cover large-scale scenarios, they lack accident scenarios crucial for a comprehensive safety assessment of autonomous driving. To address this gap, the DeepAccident dataset \cite{DeepAccidentDataset2023} was created using the CARLA simulator, featuring 57k frames and 285k annotated samples for evaluating motion and accident prediction capabilities. Data is captured by four vehicles and one infrastructure unit using multi-view RGB cameras and LiDAR, with corresponding annotations for perception and motion prediction tasks. This dataset provides safety-critical scenarios and introduces an end-to-end accident prediction task for forecasting collision details. V2XFormer, proposed as a benchmark, demonstrates superior performance in motion and accident prediction compared to single-vehicle models. However, DeepAccident lacks scenario diversity, realistic sensor models, environmental conditions, and vulnerable road users. The SCOPE dataset \cite{SCOPE} addresses these limitations as the first synthetic multi-modal dataset for CP, incorporating realistic sensor models, physically accurate weather simulations, over 40 diverse scenarios with up to 24 collaborative agents, and passive traffic. It also provides benchmarks for 2D/3D object detection and semantic segmentation.}

\section{\textcolor{black}{Performance Validation and Field Experiment}}
\label{Performance_Validation}

\textcolor{black}{
Evaluating CP systems requires a blend of simulations, offline testing, and real-world experiments. Simulations and offline validation allow for controlled performance assessments of V2X-based CP approaches, optimizing metrics to achieve optimal performance. In contrast, field experiments test CP adaptability in real-world conditions, revealing how systems handle diverse traffic and communication challenges. This section first compares the performance of various CP methods in offline testing and then explores real-world initiatives, including experimental deployments, pilot projects, and field tests, to assess CP readiness for practical deployment.
}

\subsection{\textcolor{black}{Performance Validation}}

\textcolor{black}{
In this subsection, performance validation is presented through a synthesis of experimental results from existing studies on CP methods for autonomous driving, using both real-world and simulated datasets, with 3D object detection used as an example downstream task. Average Precision (AP), defined as the area under the precision-recall curve, serves as the primary metric for 3D object detection, with APs at Intersection-over-Union (IoU) thresholds of 0.5 and 0.7 utilized to assess various models. Performance comparisons are presented across several datasets, including the real-world datasets DAIR-V2X \cite{dairv2x2022yu} and V2V4Real \cite{Xu2023V2V4RealAR}, as well as simulation-based datasets OPV2V \cite{opv2v2022xu}, V2X-Sim \cite{v2xsim2022li}, and V2XSet \cite{v2xvit2022xu}.}

\subsubsection{\textcolor{black}{Detection Performance Comparison}}
\textcolor{black}{To ensure fairness and generalizability, all reported results reflect the best outcomes from experimental evaluations of existing methods, with only those methods assessed on at least two datasets included. Table \ref{3D_detect} displays the AP values at IoU thresholds of 0.5 and 0.7 across five datasets under default settings, highlighting the highest AP values and indicating the source for each result. The observed performance differences across methods and datasets indicate that certain algorithms are more effective under specific environmental or data conditions.}

\textcolor{black}{
Table \ref{3D_detect} results indicate that DI-V2X achieves the highest AP@0.5 on DAIR-V2X at 0.7882, while DiscoNet leads on V2V4Real with an AP@0.5 of 0.7360. For AP@0.7 on V2V4Real, both FPV-RCNN and MACP achieve top performance at 0.4790, though MACP scores slightly lower than FPV-RCNN with an AP@0.5 of 0.6760. Among simulation-based datasets, MKD-Cooper performs exceptionally on V2X-Sim, with an AP@0.5 of 0.9300 and AP@0.7 of 0.8520, and achieves a high AP@0.7 of 0.9240 on OPV2V, positioning it as a leading method in simulated environments. V2X-PC demonstrates robust performance across both real and simulated datasets, achieving an AP@0.5 of 0.7689 and AP@0.7 of 0.6939 on DAIR-V2X and even higher scores on V2XSet with an AP@0.5 of 0.9283 and AP@0.7 of 0.8955. These findings suggest that V2X-PC excels across diverse datasets, indicating strong adaptability to various conditions.}

\begin{table*}[htp]
\renewcommand{\arraystretch}{1.3}
\caption{\textcolor{black}{3D Object Detection Performance Comparison On DAIR-V2X, V2V4Real, OPV2V, V2XSet, and V2X-Sim datasets, with AP@0.5 and AP@0.7.}}\label{3D_detect}
\renewcommand\tabcolsep{2.2pt}
\centering
\begin{tabular}{lccccccccccc}
\toprule
\multicolumn{1}{c}{\multirow{4}{*}{Method}} &\multicolumn{1}{c}{\multirow{4}{*}{Year}} & \multicolumn{4}{c}{Real-world Datasets}  & \multicolumn{6}{c}{Simulation-based   Datasets} \\
\cmidrule(l{1pt}r{1.5pt}){3-6} \cmidrule(l{1.5pt}r{1pt}){7-12}
& & \multicolumn{2}{c}{DAIR-V2X} & \multicolumn{2}{c}{V2V4Real} & \multicolumn{2}{c}{OPV2V} & \multicolumn{2}{c}{V2XSet} & \multicolumn{2}{c}{V2X-Sim}  \\
\cmidrule[0.05pt](l{1.5pt}r{1.5pt}){3-4} \cmidrule[0.05pt](l{1.5pt}r{1.5pt}){5-6} \cmidrule[0.05pt](l{1.5pt}r{1.5pt}){7-8} \cmidrule[0.05pt](l{1.5pt}r{1.5pt}){9-10} \cmidrule[0.05pt](l{1.5pt}r{1.5pt}){11-12}
& & AP@\(0.5/0.7\) & Source & AP@\(0.5/0.7\) & Source & AP@\(0.5/0.7\) & Source  & AP@\(0.5/0.7\) & Source & AP@\(0.5/0.7\) & Source \\ 
\midrule
F-Cooper \cite{fcooper2019chen} 
& 2019
& \(0.7370/0.5600\) & \cite{ni2024self} & \(0.6930/0.4320\) & \cite{ni2024self} & \(0.9070/0.8100\) & \cite{li2023s2r} & \(0.8400/0.6800\) & \cite{v2xvit2022xu} & \(0.7150/0.5470\) & \cite{MKD_Cooper}   \\
V2VNet \cite{v2vnet2020wang} 
& 2020 
& \(0.7097/0.4763\) & \cite{Domain_Invariant_CP_3DOD} & \(0.6470/0.3360\) & \cite{V2VFormer}  & \(0.9350/0.7400\) & \cite{robust2022lu} & \(0.8710/0.6460\) & \cite{RoCo} & \(0.8850/0.6960\) & \cite{MKD_Cooper}   \\
When2com \cite{liu2020when2com} & 2020
& \(0.5188/0.3705\) & \cite{BM2CP} & - & - & \(0.7785/0.6240\) & \cite{CP_based_on_ST_feature} & \(0.7016/0.5372\) & \cite{yang2023spatio} & \(0.5206/0.4421\) & \cite{wang2023umc}  \\
Who2com \cite{liu2020who2com} 
&2020 & - & - & - & - & \(0.5536/0.2339\)   & \cite{wang2023umc}   & - & - & \(0.4977/0.4230\) & \cite{wang2023umc}  \\
V2VNet$_{\mathrm{robust}}$   \cite{RobustV2VNet} 
& 2021
& \(0.6610/0.4860\) & \cite{ni2024self} & \(0.5500/0.3090\)   & \cite{ni2024self} & \(0.9420/0.8540\) & \cite{ni2024self}  & -  & - & \(0.8400/0.7540\) & \cite{robust2022lu} \\
DiscoNet \cite{disconet2021li} 
&2021
& \(0.7370/0.5840\) & \cite{ni2024self} & \textbf{0.7360}\(/0.4660\) & \cite{ni2024self} & \({0.9160}/0.7910\) & \cite{robust2022lu} & \(0.9078/0.8381\) & \cite{V2X-PC} & \(0.8180/0.5370\) & \cite{MKD_Cooper}   \\
OPV2V \cite{opv2v2022xu} 
& 2022& \(0.7330/0.5530\) & \cite{robust2022lu} & \(0.6450/0.3430\)& \cite{V2VFormer}  & \(0.9430/0.8270\) & \cite{robust2022lu} &\(0.9188/0.8475\)& \cite{V2X-PC} & \(0.8870/0.7340\) & \cite{MKD_Cooper}   \\
Where2comm \cite{hu2022where2comm} 
&2022& \(0.7520/0.5880\) & \cite{ni2024self} & \(0.7040/0.4690\) & \cite{ni2024self} & \(0.9440/0.8550\) & \cite{ni2024self} & \(0.9130/0.8530\) & \cite{V2X-M2C} & \(0.6205/0.5459\) & \cite{wang2023umc}  \\
V2X-Vit \cite{v2xvit2022xu} 
& 2022& \(0.7398/0.6150\) & \cite{V2X-PC} & \({0.6800}/0.3910\) & \cite{ni2024self} & \(0.9460/0.8560\) & \cite{robust2022lu} & \(0.9100/0.8030\) & \cite{RoCo} & \(0.9010/0.7950\) & \cite{MKD_Cooper}   \\
CoBEVT \cite{xu2022cobevt}  
&2022 & \(0.6390/0.5167\) & \cite{V2X-PC} & \(0.6650/0.3600\) & \cite{V2VFormer} & \(0.9140/0.8620\) & \cite{wang2023core} & \(0.9033/0.8269\) & \cite{V2X-PC}& - & - \\
FPV-RCNN \cite{keypoints2022yuan} 
&2022 & \(0.6550/0.5050\) & \cite{robust2022lu} & \(0.7010/\)\textbf{0.4790} & \cite{ni2024self} & \(0.8580/0.8400\) & \cite{robust2022lu}& \(0.8650/0.5630\) & \cite{RoCo} & \(0.8700/0.8380\) & \cite{robust2022lu} \\
UMC \cite{wang2023umc} 
&2023 & - & - & - & - & \(0.6190/0.2450\) & \cite{wang2023umc} & - & - & \(0.6780/0.6001\) & \cite{wang2023umc}  \\
What2comm   \cite{yang2023what2comm} 
&2023 & \(0.6081/0.4463\) & \cite{yang2023what2comm} & - & - & \(0.8683/{0.7361}\) & \cite{yang2023what2comm} & \(0.8459/0.6686\) & \cite{yang2023what2comm} & - & - \\
SCOPE \cite{yang2023spatio} 
&2023 & \(0.6518/0.4989\) & \cite{yang2023spatio} & - & - & \(0.8971/0.8062\) & \cite{yang2023spatio} & \(0.8752/0.7505\) & \cite{yang2023spatio} & - & - \\
BM2CP \cite{BM2CP} 
&2023 & \(0.6403/0.4899\) & \cite{BM2CP} & - & - & \(0.8372/0.6317\) & \cite{BM2CP} & - & - & - & - \\
CoAlign \cite{robust2022lu} 
& 2023 & \(0.7460/0.6040\) & \cite{robust2022lu} & \(0.7090/0.4170\) & \cite{ni2024self} & \textbf{0.9660}\(/0.9120\) & \cite{robust2022lu} & \(0.9190/0.8050\) & \cite{RoCo} & \(0.8580/0.7650\) & \cite{robust2022lu} \\
V2VFormer \cite{V2VFormer} 
&2024 & - & - & 0.6540\(/\)0.3610 & \cite{V2VFormer}  & \(0.9170/0.8760\) & \cite{V2VFormer} & - & - & 0.6150\(/\)0.5500 & \cite{V2VFormer}    \\
CoBEVGlue \cite{ni2024self} &2024 & 0.7400\(/\)0.5820 & \cite{ni2024self} & 0.7020\(/\)0.4310 & \cite{ni2024self} & 0.9580\(/\)0.9090 & \cite{ni2024self} & - & - & - & - \\
IFTR \cite{IFTR} &2024 & 0.2058\(/\)0.0665 & \cite{IFTR} & - & - & 0.8556\(/\)0.6604 & \cite{IFTR} & 0.7173\(/\)0.4967 & \cite{IFTR} & - & - \\
V2X-M2C \cite{V2X-M2C} &2024 & - & - & - & - & 0.8870\(/\)0.8280 & \cite{V2X-M2C} & 0.9220\(/\)0.8600 & \cite{V2X-M2C} & - & - \\
CPSC   \cite{CP_based_on_ST_feature} 
&2024 &0.6542\(/\)0.4784 & \cite{CP_based_on_ST_feature} & - & - & 0.8911\(/\)0.7797 & \cite{CP_based_on_ST_feature} & - & - & - & - \\
V2X-INCOP \cite{V2X-INCOP2023} 
&2024 & 0.6548\(/\)0.5526 & \cite{V2X-INCOP2023} & - & - & 0.9256\(/\)0.8318 & \cite{V2X-INCOP2023} & - & - & - & - \\
RoCo\cite{RoCo} 
&2024 & 0.7630\(/\)0.6200 & \cite{RoCo} & - & - & - & - & 0.9190\(/\)0.8050 & \cite{RoCo} & - & - \\
MACP \cite{MACP}  
&2024 & - & - & 0.6760\(/\)\textbf{0.4790} & \cite{MACP} & 0.9370/0.9030 & \cite{MACP} & - & - & - & - \\
DI-V2X   \cite{Domain_Invariant_CP_3DOD}  
&2024 &\textbf{0.7882}\(/\)0.6616 & \cite{Domain_Invariant_CP_3DOD} & - & - & - & - & 0.9270\(/\)0.8270 & \cite{Domain_Invariant_CP_3DOD} & - & - \\
MKD-Cooper \cite{MKD_Cooper} 
&2024 & - & - & - & - & \(0.9650/\){\textbf{0.9240}} & \cite{MKD_Cooper} & - & - & \textbf{0.9300}\(/\)\textbf{0.8520} & \cite{MKD_Cooper}   \\
V2X-PC \cite{V2X-PC} &2024 & 0.7689\(/\)\textbf{0.6939} & \cite{V2X-PC} & - & - & - & - & \textbf{0.9283}\(/\)\textbf{0.8955} & \cite{V2X-PC} & - & - \\ 
\bottomrule
\end{tabular}
\end{table*}

\subsubsection{\textcolor{black}{Robustness to Time Delay}}

\begin{figure}
    \centering
    \includegraphics[width=\linewidth]{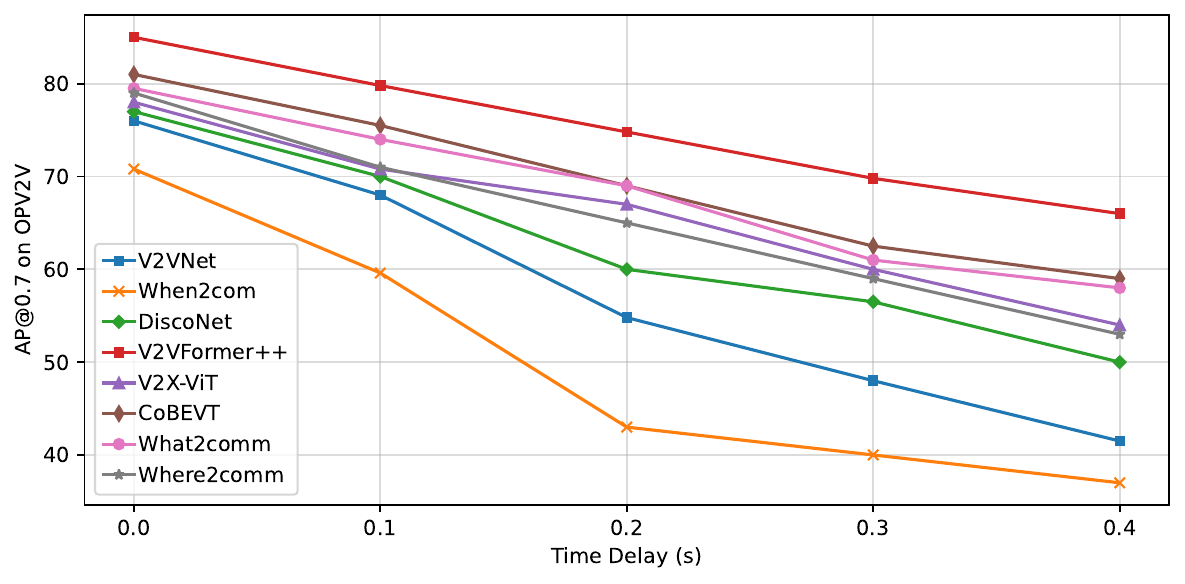}
    \caption{The impact of time delay on the AP@0.7 performance metric for the OPV2V dataset. The results are referenced from \cite{yang2023what2comm} and \cite{V2VFormer++}.}
    \label{time_delay}
\end{figure}

\textcolor{black}{
Time delay introduced by V2X communication presents a significant challenge in CP, leading to data asynchronization between ego and other agents during the data fusion process. To illustrate model robustness to time latency, we present performance across delays ranging from 0 to 0.4 seconds, as shown in Fig. \ref{time_delay}, with results referenced from \cite{yang2023what2comm} and \cite{V2VFormer++}.}

\textcolor{black}{
As shown in Fig. \ref{time_delay}, detection performance declines for all methods as time delay increases. V2VFormer++ maintains relatively high AP scores, demonstrating the greatest resilience to latency. CoBEVT and What2comm also show robust performance, though to a lesser extent. Where2com and V2X-ViT exhibit moderate robustness, with gradual AP declines indicating some sensitivity to delay. In contrast, DiscoNet and V2VNet experience sharper AP declines, reflecting higher latency sensitivity. When2com begins with lower AP scores and degrades significantly as delay increases, highlighting its vulnerability to latency. In summary, V2VFormer++ proves the most delay-tolerant, while When2com is most affected by latency in the OPV2V setting, limiting its effectiveness under substantial transmission delays.}

\subsubsection{\textcolor{black}{Robustness to Pose Errors}}

\begin{figure}
    \centering
    \includegraphics[width=\linewidth]{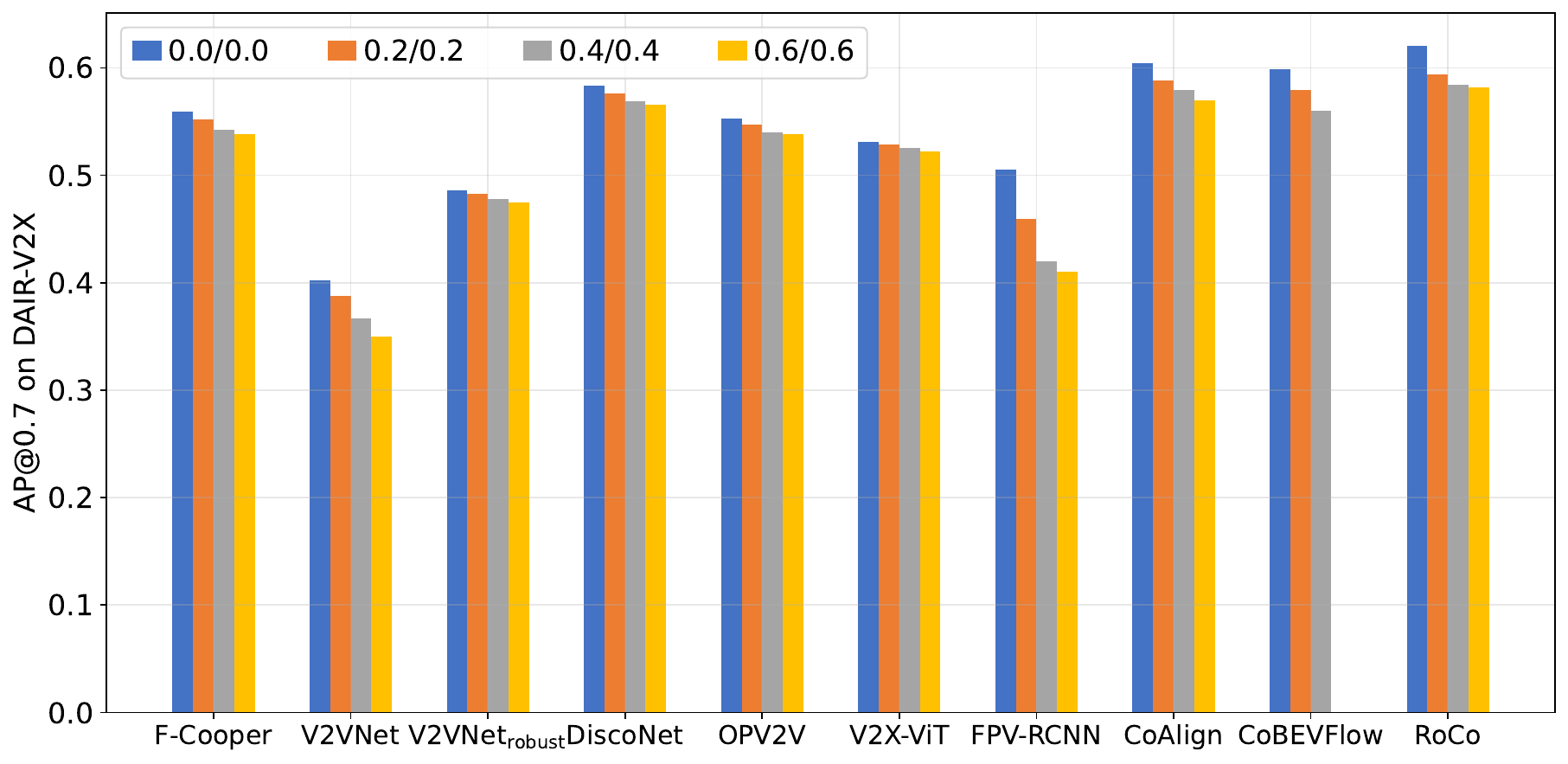}
    \caption{Detection performance on DAIR-V2X dataset with pose errors at four different noise levels. The pose noises follow Gaussian distribution \(\mathcal{N}(0,\sigma_t)\) on \(x, y\) and \(\mathcal{N}(0,\sigma_r)\) on \(\theta\), where \(x, y, \theta\) are 2D centers and yaw angle of accurate global poses. We present AP@0.7 at noise level \(\sigma_t/\sigma_r=0.0m/0.0^\circ, 0.2m/0.2^\circ, 0.4m/0.4^\circ, 0.6m/0.6^\circ\). The results are referenced from \cite{robust2022lu} and \cite{RoCo}.}
    \label{pose_noise}
\end{figure}

\textcolor{black}{
Collaborative agents rely on precise pose data from others to accurately transform the coordinates of the data in the received messages. However, even with advanced localization technologies such as GPS, pose errors remain unavoidable, requiring collaborative approaches to be resilient to these inaccuracies. Following \cite{robust2022lu}, Gaussian noise \(\mathcal{N}(0,\sigma_t)\) is added to \(x, y\), and \(\mathcal{N}(0,\sigma_r)\) to \(\theta\), to simulate pose errors, where \(x, y, \theta\) represent the 2D center and yaw angle of accurate global poses. Fig. \ref{pose_noise} shows the detection performance in terms of AP@0.7 for ten different methods on the DAIR-V2X dataset, evaluated at four levels of pose error \((0.0m/0.0^\circ, 0.2m/0.2^\circ, 0.4m/0.4^\circ, 0.6m/0.6^\circ)\), with results sourced from \cite{robust2022lu} and \cite{RoCo}.}

\textcolor{black}{
Among these methods, RoCo maintains the highest AP scores across noise levels, demonstrating strong resilience to pose errors. In contrast, V2VNet shows the lowest AP with substantial declines at higher noise levels, while its robust variant, V2VNet$_{\mathrm{robust}}$, outperforms the standard version, indicating improved robustness to pose noise. CoAlign and CoBEVFlow perform well but yield slightly lower AP than RoCo. F-Cooper, DiscoNet, OPV2V, and V2X-ViT display stable performance across noise levels, suggesting greater resilience to pose errors. FPV-RCNN, despite moderate detection accuracy, shows notable AP declines as noise increases, indicating higher sensitivity to positional errors. Overall, increasing pose errors reduce AP@0.7 across all methods, underscoring their sensitivity to positional inaccuracies.}

\subsection{\textcolor{black}{Field Experiments and Pilot Projects}}

\textcolor{black}{
Field experiments of CP systems have progressed through a range of global pilot projects, demonstrating CP's adaptability and effectiveness under specific conditions. These initiatives showcase CP’s potential to improve traffic efficiency, safety, and the functionality of CAVs within controlled environments, advancing research and development in intelligent transportation systems.}

\textcolor{black}{
The Cooperative Adaptive Cruise Control (CACC) project carried out in the U.S. leverages DSRC-based communication to enable real-time sharing of speed and positioning data between vehicles, allowing for autonomous maintenance of safe inter-vehicle distances \cite{CACC}. The project aims to enhance Adaptive Cruise Control (ACC) by dynamically coordinating vehicle spacing within a string, ultimately improving traffic flow. Test cases for this project include simultaneous passing of vehicles on adjacent lanes, driving in a string with induced oscillations to assess string stability, handling communication interruptions in CACC mode, and evaluating the effects of positional inaccuracies in preceding or adjacent vehicles. }

\textcolor{black}{
The Autonomous Intersection Management (AIM) project carried out in the U.S. explores autonomous traffic management at intersections without traditional traffic signals, using a scalable multi-agent framework suitable for urban areas \cite{AIM}. Recognizing the coexistence of autonomous and human-driven vehicles, AIM introduced the “Hybrid AIM” protocol to facilitate safer, more efficient integration at intersections. The project employs augmented reality to create a mixed-reality setup, simulating virtual traffic on real roads, allowing for comprehensive testing of CP-enabled systems in dynamic and complex scenarios. This project offers a valuable template for urban areas aiming to reduce congestion through CP, highlighting how CP can enhance intersection management by enabling seamless coordination among diverse vehicle types.}

\textcolor{black}{
The Ko-PER Project, carried out in Germany, focuses on road safety, using distributed sensor networks to build a comprehensive view of local traffic \cite{6856479}. By integrating data from vehicle-mounted and infrastructure sensors, Ko-PER enables the detection of occluded traffic participants and the evaluation of collision risk \cite{KoPER}. A dataset has been produced from this project, containing laser scanner and video data from public intersections, and supports ongoing research on intersection perception and cooperative warning systems \cite{6957976}.}

The Virginia Connected Corridors (VCC) project in the U.S. is led by the Virginia Department of Transportation and the Virginia Tech Transportation Institute \cite{VCC}. It serves as a testbed for CAV applications. Since its launch in 2016, a fleet of 50 highly instrumented light vehicles and 60 Roadside Units (RSUs) have been deployed within the Northern Virginia Test Bed for application testing.
Each vehicle is equipped to communicate via DSRC with the surrounding connected infrastructure. When the vehicles are outside the range of a DSRC-equipped RSU, they can use cellular technology for communication. A centralized cloud system is employed to manage the message traffic between connected vehicles. The vehicles can connect to the cloud through either DSRC or cellular networks. 
It has been noted that cellular communications have a longer latency, ranging from 1.5 to 3.5 seconds, compared to DSRC and are therefore used for applications where timely message receipt is not critical. The aim of this project is to provide an open development platform, allowing third-party developers to test and deploy applications that enhance roadway safety and mobility.

The iMOVE Cooperative Perception Project, led by the Australian Centre for Field Robotics at the University of Sydney in collaboration with Cohda Wireless, has demonstrated advancements in pedestrian detection around corners through the use of CPMs and Dedicated Short-Range Communication (DSRC) technology \cite{imove_j}. Notably, this project illustrates how CAVs can autonomously and safely interact with pedestrians, both walking and running, relying exclusively on CP information provided by an intelligent RSU \cite{imove_j}. This project represents one of the pioneering demonstrations of urban vehicle automation that relies solely on CP data for safe and efficient operation in complex environments.

These projects demonstrate the versatility of CP across various traffic scenarios, including highway platooning, intersection management, and pedestrian detection in urban areas. They underscore CP's important role in addressing perception gaps caused by occlusions or limited sensor range, ultimately leading to more reliable and robust autonomous driving systems. Additionally, the challenges faced during these projects will inform the development of next-generation CP algorithms and infrastructure to create safer and more efficient autonomous transportation ecosystems.

\section{Lessons Learned} \label{challenges}

CP technology plays a pivotal role in overcoming the limitations inherent in single autonomous vehicles, addressing issues like restricted perception range and obscured areas. However, the unique characteristics of autonomous driving and the evolution of communication technology bring forth a multitude of challenges for CP. Drawing from recent research and comprehensive analysis of cooperative autonomous driving, we have gleaned valuable insights and identified open challenges and future directions that will significantly influence the research trajectory in this field.

\subsection{Challenges and Potential Solutions}

\subsubsection{Collaboration Trigger}
A prevailing challenge within CP is the ability of each CAV to autonomously identify optimal opportunities for collaboration. Current research frequently emphasizes predefined cooperative behaviors, often neglecting the necessity for ego CAVs to dynamically evaluate when collaboration is advantageous. This process necessitates the development of a mechanism referred to as a ``collaboration trigger," which assesses the trade-offs between independent actions and collaborative efforts, informed by real-time conditions such as weather, road attributes, and sensor performance. Furthermore, ensuring stable and temporally consistent cooperation among multiple CAVs presents significant challenges, particularly in the context of dynamic driving scenarios.

\textcolor{black}{
To address these challenges, adaptive, context-aware systems are needed to determine when collaboration is optimal. A dynamic decision model allows each ego CAV to assess real-time factors like sensor performance, road conditions, and traffic, ensuring stable and efficient cooperation. Key metrics, such as perception accuracy and energy efficiency, can guide these decisions, with a fusion algorithm dynamically weighing factors like weather and traffic forecasts as conditions change. Effective implementation requires a cloud or edge-computing platform to aggregate real-time and historical data, enabling CAVs to make informed, anticipatory collaboration choices. 
}

\subsubsection{Real-world Communication Constraints}
Existing CP methods have traditionally focused on optimizing perception performance, often oversimplifying the critical challenges posed by real-world communication constraints such as limited bandwidth, unstable channels, network congestion, and data delays. These factors compromise the data received by the ego vehicle, and unlike conventional applications, retransmitting data in CP can render it unusable due to outdated information \cite{liu2023cooperative}. This results in missing data, asynchronous information from other agents, and, ultimately, inaccurate data fusion within the CP system.

\textcolor{black}{
A promising approach is the integration of cross-layer models to assess how communication constraints affect CP algorithms \cite{Survey_Integrated_Simulation_Environments_CP}. For example, combining network-layer routing with physical-layer channel modeling enables CP systems to anticipate the effects of channel fading or congestion on data transmission, facilitating adaptive adjustments in CP processes. Additionally, predictive perception models can enhance CP resilience by using historical data and probabilistic modeling to forecast surrounding vehicle behavior, maintaining situational awareness even during data loss. This reduces dependency on continuous communication, ensuring reliable CP performance in real-world conditions.}

\subsubsection{Malicious and Selfish Behavior}
Trustworthy communication is essential for fostering effective and safe CP. However, some road users may act in self-serving ways, collaborating only to reduce their own costs while potentially harming others. These self-interested users, commonly known as ``free-riders,” exploit shared information to benefit from coordinated efforts without contributing their own insights to the collective knowledge pool. Furthermore, the distributed deployment of multiple agents lowers defenses against cyber-attacks, increasing vulnerability to malicious activities by untrustworthy users \cite{han2023secure}. Malicious collaborators may deliberately transmit inaccurate environmental data, compromising the integrity of CP by degrading data quality and reliability. This degradation not only disrupts CP effectiveness but also heightens risks to the safety of CAVs that rely on accurate, shared situational awareness.

To tackle the challenges posed by unreliable collaborators in CP, several strategies can enhance data reliability and safety. Implementing trust scoring and reputation systems allows CAVs to prioritize data from agents with a history of reliability. Machine learning-based anomaly detection can also identify unusual patterns that might indicate malicious behavior. Cross-verifying data from multiple sources and utilizing multi-modal sensor fusion further improves the accuracy of CP, enabling CAVs to detect and disregard inconsistent or unreliable inputs. Blockchain technology can introduce a decentralized layer of trust by immutably logging data interactions, increasing agent accountability. Lastly, dynamic data prioritization ensures that critical information, such as immediate obstacles, is prioritized, reducing the potential impact of harmful data. Together, these approaches strengthen the resilience of cooperative perception, supporting safer and more reliable collaborative driving.

\subsubsection{Generalizability in Real Scenarios}
Constructing real-world scenes and collecting data for CP research presents substantial challenges, leading many CP methods to rely on simulated datasets for development. While some recent real-world datasets have emerged \cite{dairv2x2022yu, v2x-seq, HoloVIC, Xu2023V2V4RealAR, axmann2023lucoop, TUMTrafV2X}, they typically involve limited agent participation, which fails to capture the full complexity of real-world collaborative scenarios. As a result, CP frameworks trained primarily on simulated datasets may struggle to generalize effectively in realistic collaboration settings.

To address this gap, domain adaptation and synthetic data enhancement techniques offer promising solutions that can improve the applicability of CP models to real-world environments. Domain adaptation models, for instance, align features between simulated and real-world datasets through adversarial training and transfer learning, reducing distributional discrepancies and enabling CP models to perform more reliably in diverse, real-world settings \cite{wilson2020survey}.

\textcolor{black}{
In addition to domain adaptation, synthetic data enhancement techniques improve the realism of simulated datasets. One approach, described in \cite{DataSynthesisPipeline}, synthesizes training data using Augmented Reality, then refines it through a Generative Adversarial Network (GAN) to create robust photo-realistic images under various weather and lighting conditions. Other generative models, such as diffusion models, also play a vital role by transforming simulated data into realistic visuals. These models adjust textures, lighting, and environmental effects, effectively bridging the gap between synthetic and real-world visuals. By enhancing dataset realism, these techniques support more robust CP model training and improve generalization across varied real-world scenarios.}

\subsubsection{Challenging Scenes and Corner Cases}
While several large-scale CP datasets have been developed in recent years, many focus primarily on typical traffic situations and often neglect challenging scenarios, particularly corner cases. Corner cases are uncommon and extreme situations in autonomous driving, characterized by unusual conditions, hardware limitations, or irregular obstacles, such as overturned vehicles or unusually shaped objects. These rare scenarios are frequently excluded from perception model training, which hinders the model’s ability to identify objects accurately and can lead to missed or erroneous detections. Such inaccuracies in perception increase the risk of severe traffic accidents, highlighting the urgent need for CP datasets that incorporate both intricate scenarios and corner cases to support the development of more robust perception systems.

\textcolor{black}{
A combined approach of targeted data collection and synthetic data generation is essential to address the gap in corner cases within CP datasets. Simulation environments can be utilized to create synthetic data that replicates corner cases, including overturned vehicles, unusual object shapes, and scenarios with sensor occlusions. This method allows for the generation of a wide variety of challenging conditions without relying on rare real-life occurrences. Additionally, a model-driven data collection approach can be implemented, where trained CP models actively monitor their predictions in real-time. They can identify and flag low-confidence detections or false positives/negatives as potential corner cases. These flagged instances can then be collected and prioritized for further training. By creating datasets that encompass both typical and rare autonomous driving conditions, CP systems can be better trained and validated, ultimately improving performance and reliability in real-world situations.}

\subsection{Future Directions}
\label{future_direction_and_outlook}

\subsubsection{Integrated Sensing and Communication for CP}
Integrated Sensing and Communication (ISAC) is an emerging technology that utilizes the same frequency band and hardware for both sensing and communication \cite{10012421}. This innovation has the potential to significantly improve spectrum efficiency and reduce hardware costs. In the current context of cooperative autonomous driving, sensing and communication systems typically operate on separate frequency bands. The widespread adoption of millimetre-wave and massive MIMO technologies in infrastructure opens opportunities to implement radar perception in cooperative systems through ISAC. This enhances wireless communication throughput and radar perception resolution in future V2X networks. ISAC is anticipated to be a critical technology in future V2X CP, enabling low-latency, high-data-rate communication, and high-resolution obstacle detection, particularly in challenging scenarios like high occlusion and adverse weather.

\subsubsection{Responsible AI for CP}
Cooperative autonomous driving relies heavily on Artificial Intelligence (AI), which excels in mastering increasingly complex computational tasks. 
Responsible AI initiatives can offer valuable insights for creating realistic simulation environments that closely resemble real-world scenarios, benefiting the training and validation of V2X CP frameworks. Additionally, these initiatives can drive the development of domain adaptation techniques, enhancing knowledge transfer between simulated and real domains. This knowledge transfer improves the generalisation of the CP model. Moreover, responsible AI efforts can stimulate the advancement of deep learning methods tailored for generating CP data featuring intricate scenarios and corner cases. Generative models and ethically guided data augmentation techniques are pivotal for constructing comprehensive datasets. This enriched dataset forms the basis for robust training for the V2X CP system, equipping them to navigate and respond effectively to challenging real-world situations.

\subsubsection{Privacy-preserved CP}
To establish sustainable CP, ensuring the privacy of data and messages exchanged via V2X communications is of paramount importance \cite{9845490}. 
FL offers a viable solution to enhance CP intelligence while preserving privacy. FL facilitates intelligent CP by enabling fusion algorithms to run on servers without requiring access to raw data or information sharing. It employs a distributed training mechanism in which collaborating vehicles train localized CP models using their data and then share only the trained model with the server for global aggregation \cite{barbieri2022decentralized}. This approach ensures secure distributed learning, safeguarding the shared model from external adversaries. By prioritizing privacy preservation, FL encourages more CAVs to participate in data perception, thus enhancing the overall robustness of CP systems.

\subsubsection{Collaborative AI Integration for Enhanced CP and V2X Communication}
In future intelligent transportation systems, efficient collaboration is envisaged between AI systems for V2X communication and AI systems for CP within each CAV.

The integration of AI into radio resource management for vehicle communication systems is driven by the imperative to allocate communication resources efficiently, enabling real-time data exchange among vehicles, infrastructure, and relevant entities \cite{9214878}. International standardization efforts, led by organizations like the 3GPP, are underway in anticipation of the 6G era. The forthcoming 6G technologies promise unprecedented capabilities in data rates, low latency, and accommodating a massive number of connected devices, all of which are crucial for CP in automated and connected transportation systems.

In the context of CP, each CAV must intelligently determine the information it requires from other connected vehicles and decide what information it should transmit to meet requests from other vehicles. This decision-making process is highly contingent on the dynamic traffic conditions. Therefore, each CAV's AI system must efficiently manage CP's necessary information, adapting to the current and near-future traffic environment. However, to optimize information flow management, each CAV must collectively consider the available communication resources at its disposal.

Consequently, further research is imperative to orchestrate effective collaboration and coordination among distinct AI systems operating within individual CAVs and AI systems operating within the V2X network infrastructure. These AI systems must align their objectives and strategies to achieve efficient CP, particularly in dynamic and complex traffic scenarios. Such endeavours are essential to harnessing the full potential of AI-driven CP to enhance road safety and traffic efficiency.

\subsubsection{Migration of Advanced Perception Framework from Single Vehicle to Cooperative Vehicles} 
\textcolor{black}{As the development of transformer models in both computer vision and large language model areas, several advanced technologies have been proposed for single-vehicle perception and autonomous driving, e.g., end-to-end perception \cite{DQTrack}, 3D occupancy grid prediction \cite{Scene_as_occupancy}, referring expression-based perception \cite{referring_expression_based_perception} with multi-modal large language model \cite{LiDARLLM}, self-supervised world-models for autonomous driving to boost the downstream tasks by using the features extracted from understanding and predicting/reconstruction of the environment, and end-to-end driving. All of the aforementioned technologies have great potential to be extended into cooperative vehicle settings as well, with the expectation of enhancing the capability of autonomous driving. The most recent examples are \cite{CP_E2E, SimulationPlatformCooperativeEndtoEndAD, CollaborativeOccupancyPrediction_tiv, CollaborativeOccupancyPrediction}, in which the end-to-end autonomous driving framework and 3D occupancy grid prediction function are integrated with V2X cooperation, respectively, to achieve significant performance improvement on planning and intermediate perception outputs.}

\section{Conclusions}\label{conclusions}

In this paper, we presented a comprehensive framework for V2X CP, addressing its foundational components and key aspects. Through a novel taxonomic classification of mainstream architectures and an in-depth literature review, we analyzed the multifaceted challenges inherent to effective CP. We examined how V2X communication limitations, such as latency and data loss, can impact CP performance, emphasizing the need for resilient communication strategies to maintain reliable perception sharing. We also highlighted the value of open-source databases and simulators, which serve as essential tools for evaluating and refining CP models in controlled environments. These resources enable rigorous testing and iterative optimization of CP algorithms, promoting advancements in model accuracy and robustness. Furthermore, we underscored the importance of embedding ethical principles, such as fairness, transparency, and privacy, into the development and deployment of AI models for CP. Incorporating these principles fosters ethical AI usage in cooperative driving and supports the creation of realistic simulation environments, domain adaptation techniques, and diverse datasets, which are crucial for equipping CP models to handle the complexities of real-world scenarios. In conclusion, CP is a rapidly evolving field with significant potential. By combining advanced V2X communication technologies, multimodal sensing, and AI-driven data processing, CP offers a pathway toward reliable, efficient, and ethically sound solutions for autonomous driving and intelligent transportation systems, ultimately enabling safer, more effective, and highly capable autonomous mobility.

\footnotesize
\bibliographystyle{IEEEtran}
\bibliography{mybibfile}

\end{document}